\documentclass[journal]{IEEEtran}
\pdfoutput=1
\ifCLASSOPTIONcompsoc
\usepackage[tight,normalsize,sf,SF]{subfigure}
\else
\usepackage[tight,footnotesize]{subfigure}
\fi
\usepackage{color}
\ifCLASSINFOpdf
\else
\fi

\usepackage{graphicx}
\usepackage{float}
\usepackage[intlimits]{amsmath}
\usepackage[lined, ruled,commentsnumbered]{algorithm2e}
\usepackage{setspace}
\usepackage{multirow}
\usepackage{cite}
\usepackage{mathptmx}
\usepackage{amsfonts}
\usepackage{stmaryrd}
\usepackage{color}
\usepackage{booktabs}
\usepackage{multirow}
\usepackage{algorithmic}
\graphicspath{ {Comparative_studies/image/} }

\begin{document}

%
\title{Explainable and Safe Reinforcement Learning for Autonomous Air Mobility}
%
%
%

\author{Lei Wang, Hongyu Yang, Yi Lin~\IEEEmembership{Member,~IEEE}, Suwan Yin, Yuankai Wu*~\IEEEmembership{Member,~IEEE}
        

\thanks{The authors are with the College of Computer Science, Sichuan University, Chengdu 610065, China (Corresponding author: Yuankai Wu. e-mail: wuyk0@scu.edu.cn)}}

%
%

\markboth{IEEE Transactions on Intelligent Transportation Systems}%
{Shell \MakeLowercase{\textit{et al.}}: Bare Demo of IEEEtran.cls for Journals}
%



\maketitle

\begin{abstract}
Increasing traffic demands, higher levels of automation, and communication enhancements provide novel design opportunities for future air traffic controllers (ATCs). This article presents a novel deep reinforcement learning (DRL) controller to aid conflict resolution for autonomous free flight. Although DRL has achieved important advancements in this field, the existing works pay little attention to the explainability and safety issues related to DRL controllers, particularly the safety under adversarial attacks. To address those two issues, we design a fully explainable DRL framework wherein we: \romannumeral1) decompose the coupled Q value learning model into a safety-awareness and efficiency (reach the target) one; and \romannumeral2) use information from surrounding intruders as inputs, eliminating the needs of central controllers. In our simulated experiments, we show that by decoupling the safety-awareness and efficiency, we can exceed performance on free flight control tasks while dramatically
improving explainability on practical. In addition, the safety Q learning module provides rich information about the safety situation of environments. To study the safety under adversarial attacks, we additionally propose an adversarial attack strategy that can impose both safety-oriented and efficiency-oriented attacks. The adversarial aims to minimize safety/efficiency by only attacking the agent at a few time steps. In the experiments, our attack strategy increases as many collisions as the uniform attack (i.e., attacking at every time step) by only attacking the agent four times less often, which provide
insight into the capabilities and restrictions of the DRL in future ATC designs. The source code is
publicly available at https://github.com/WLeiiiii/Gym-ATC-Attack-Project.
\end{abstract}


%
\IEEEpeerreviewmaketitle

\section{Introduction}

\IEEEPARstart{T}he global air movement has grown enormously in recent years, e.g., it is predicted that the number of aircraft handled by en-route centers will be increased at an average rate of 1.5\% per year from 2020 to 2040 \cite{federal2020faa}. Especially low-altitude airspace contained with relatively small aircraft like drones and helicopters is mostly uncontrolled, while predictions are that the aircraft volume in this airspace will constantly increase\cite{balakrishnan2018blueprint}. The strong growth of air traffic volume poses an incredible challenge for the safety and the en–route sector capacity. Since air traffic controllers' human cognition will not match the high traffic complexity and density, implementing free flight and autonomous air traffic controllers (ATCs) are seen as crucial tools for solving those issues. 

Free flight is a developing ATC system without centralized control and ATC operators. With the aid of communication systems and artificial intelligence, the individual pilot can make more flight path decisions freely. As in most complex systems, distributed and cooperative decision-making is believed to be more efficient than centralized control. Several theoretical studies have shown that free flight can improve safety \cite{hoekstra2002designing}, reduce fuel consumption \cite{valenti2001cost} and bring time efficiency \cite{krozel1997conflict}. Communication, navigation, and surveillance (CNS), along with autonomous ATC, are the areas that need to be addressed to make the free flight a reality. Accordingly, autonomous conflict detection and resolution tools are the primary factors that will provide the means of achieving free flight \cite{yang2016multi}.

The development of autonomous conflict detection and resolution tools for free flights is a challenging subject. The critical challenge here is to design a system to provide real-time advisories to aircraft to ensure safe separation and handle uncertainty in real-time. The conflict resolution (CR) problem can be naturally represented as an optimization problem. Thus several optimization frameworks have been developed. The optimization algorithms for CR can be briefly divided into three categories: The first category is the swarm intelligence optimization algorithms (such as genetic algorithms \cite{durand1996automatic, guan2014strategic, cecen2019two, han2019research}, or ant colony algorithms \cite{durand2009ant}). These algorithms usually divide the CR process into discrete segments with equal times or distance lengths and optimize them separately. The algorithms of the second category \cite{menon1999optimal, liu2017large} are primarily used to design the conflict avoidance path of aircraft with optimal control theory on a single or multidimensional level. The third category is hybrid system models, which delegate the CR function of the controllers to each aircraft \cite{pappas1996conflict, tomlin1996hybrid}. The core concept is to propose decentralized management schemes that will switch the hybrid control system to different optimal control solutions according to the change in the information structure between agents. Optimization-based approaches can theoretically provide the optimal solution for CR problems. However, the optimization-based approaches are often computationally expensive or require prior knowledge, limiting their applications to reality.

The impressive ability of deep reinforcement learning (DRL) to handle challenging uncertain control problems \cite{mnih2015human}, along with the advancement in other concomitant technologies (such as wireless communications), provides a potential solution for autonomous CR tools. In DRL framework, the expensive
operation of modeling the complex interactions between aircraft is learned in an offline training step, whereas the learned policy can
be queried quickly online. There are several studies in the literature associated with DRL based ATCs. Impressive results are reported in several tasks, including CR\cite{brittain2018autonomous, pham2019machine} and preventing loss of separation\cite{bertram2020distributed, brittain2021one}. It is assumed that DRL will have a predominant role in building the control system of future ATCs. 

Although DRL technology has shown excellent performance in many ATC tasks, two challenges linger while we use DRL for real-world applications. The first challenge is the safety of the ATC systems. Despite the development of different configurations of ATC controllers, they are still vulnerable to various security issues, and various attack solutions can be exploited\cite{khan2020uav}. Seriously, the threat is getting worse with the deployment of DRL agents. Modern DRL models use deep neural networks (DNNs) as function approximators. However, DNNs have been discovered to be vulnerable to adversarial examples \cite{szegedy2013intriguing}, carefully crafted, quasi-imperceptible perturbations, which fool a DNN when added to its inputs. Moreover, adversarial examples also appear in the real world without any attacker or maliciously selected noise involved \cite{athalye2018synthesizing}. For example, a few small stickers on the ground could cause a self-driving car to move into the opposite traffic lane \cite{tencent2019experimental}. Suppose adversarial attacks could be successfully applied to real-world ATC systems. In that case, attackers could easily cause accidents and jeopardize personal safety. Therefore, it is necessary to study adversarial attack and defense mechanisms for DRL agents in ATCs.   

Another issue is the explainability. The decision-making process of DRL technology is opaque, and the lack of transparency makes it impossible for pilots and controllers to understand the internal working mode of them. The "black box" nature of the DRL model may hinder users from believing in the predicted results, especially when the model is used to make key decisions\cite{kao2018context, kiran2021deep}, because the predicted consequences may be disastrous if they act according to blind beliefs. An explainable DRL model can provide explanations for the specific action made by DRL model and help users understand its internal mechanism.

This work aims to introduce a safe and explainable DRL solution for free flight control. Our work is grounded in three observations: 1) Several ATC tasks can be naturally represented as an optimization problem with safety limits. Combining optimization cost and safety limits in one reward signal for the DRL system ignores the rich situational information of airspace, thereby limiting the learning capacity of the DRL model. Thus, we decompose the coupled learning model into two separate models: goal learning and safety learning. 2) Various real-world group behaviors are the aggregate result of the actions of individual animals, each acting according to its local perception of the dynamic environment. To imitate this behavioral "control structure",  we use a local state representation scheme, in which the free flight only requires the information of its nearest neighbors. We eliminate the need for a centralized sensing module. 3) Since the coupled learning task can be decomposed, the adversarial attacks on our target can also be decomposed into goal-oriented and safety-oriented attacks. Therefore, we further propose an efficient adversarial attack scheme, which can impose specialized adversarial attack. 

\textbf{Contributions}. In summary, the main contributions of this work are as follows:
\begin{itemize}
\item We introduce a simple yet powerful DRL framework named safety-aware deep Q networks (SafeDQN), in which two separate DQNs cooperatively learn the safety limits and optimization costs.
\item We design a local state representation scheme for the DRL agent controlling the free flight and show that local perception is better than global perception on the free flight control task.    
\item Our work is the first to study the adversarial vulnerability of DRL-based ATC controllers, and we design a time-efficient attack framework that can achieve both safety-oriented and goal-oriented attacks.
\item We conduct extensive experiments to evaluate our SafeDQN model. The experimental results show that our model is safe and highly explainable.
\end{itemize}
  
The rest of this paper is organized as follows. First, we review some related works of DRL for ATCs and adversarial attacks on DRL in Section~\ref{sec:rel}. Some preliminaries are provided in Section~\ref{sec:pre}. We then introduce the proposed SafeDQN in Section~\ref{sec:meh}. Extensive experiments and discussions are conducted in Section~\ref{sec:exp}. Finally, we conclude this paper in Section~\ref{sec:con}.



\section{Related Works}
\label{sec:rel}
\subsection{Deep Reinforcement Learning based Air Traffic Controllers}

Deep neural networks have been prevailing in reinforcement learning in recent years. We have witnessed major breakthroughs like AlphaGo\cite{silver2016mastering} defeating Lee Sedol and OpenAI Five's outstanding performance against professional dota2 team OG \cite{berner2019dota}. Inspired by those breakthroughs, the idea of applying DRL to another form of a game is generated to plan and control the aircraft's flight route in the airspace, that is, air traffic control.

There have been several attempts to apply DRL to ATC tasks. DRL for ATC and CR was first introduced in \cite{brittain2018autonomous}, where an AI agent was designed to mitigate conflicts and minimize the delay of aircraft reaching their metering fixes. Later, Pham et al. \cite{pham2019machine} demonstrated that an AI agent can effectively resolve randomly generated conflict scenarios between a pair of aircraft through vectoring maneuvers. Tran et al. \cite{tran2020interactive} developed an interactive conflict solver using reinforcement learning that leveraged human resolution maneuvers. This resulted in AI-recommended maneuvers that closely aligned with human ATC behavior. Ribeiro et al. \cite{ribeiro2020improvement} recently proposed a hybrid geometric-reinforcement learning algorithm for resolving conflicts in low-altitude airspace. While these approaches are effective for low-density airspace environments, their centralized architectures fail to handle high-density airspace as the number of intruder aircraft increases. 

Several multi-agent frameworks were recently proposed for handling high-density airspace environments. Zhao et al. \cite{zhao2021physics} proposed a physics-informed deep reinforcement learning algorithm for resolving conflicts with coordination rules in traditional airspace. In \cite{yang2020scalable}, a message-passing-based decentralized computational guidance algorithm using multi-agent Monte Carlo Tree Search (MCTS) was proposed to prevent loss of separation (LOS) for UAS in an urban air mobility (UAM) setting. A computationally efficient MDP-based decentralized algorithm was proposed in \cite{bertram2020distributed}, capable of preventing LOS for UAS in unstructured airspace. Mollinga et al. \cite{mollinga2020autonomous}  proposed a graph neural network approach to CR in free airspace by representing each aircraft as a node in a graph to handle scalability. In \cite{brittain2019autonomous}, \cite{brittain2021one}, \cite{brittain2020deep}, it is shown how a decentralized separation assurance framework can prevent LOS in high-density stochastic sectors by leveraging long short-term memory networks (LSTM) and attention, but this formulation only holds when all agents are homogeneous, or optimizing the same reward function.

So far, most work on DRL-based ATCs has studied their effect on the efficiencies of specific tasks. However, those works pay little attention to the explainability and safety of DRL algorithms. Our work is the first to study the explainability and adversarial vulnerability of a DRL agent exposed to adversarial examples.

\subsection{Adversarial Attacks against Deep Reinforcement Learning}

Adversarial examples\cite{szegedy2013intriguing} are inputting to machine learning models that an attacker has intentionally designed to cause the model to make a mistake; they are like optical illusions for machines. Adversarial examples can also manipulate DRL agents. Huang et al. \cite{huang2017adversarial} firstly attempted to attack DRL models by adding perturbations at every timestamp. Behzadan et al. \cite{behzadan2017vulnerability} showed that widely-used RL algorithms, such as DQN\cite{mnih2013playing}, TRPO\cite{schulman2015trust}, and A3C\cite{mnih2016asynchronous}, are vulnerable to adversarial inputs. The adversarial inputs can lead to degraded performance even in the presence of perturbations too subtle to be perceived by a human.
Pattanaik et al. \cite{pattanaik2017robust}generated adversarial perturbations to mislead agents to take the worst action. Another critical facet of attacking DRL models is how to select appropriate attacking moments \cite{sun2020stealthy}.
Instead of attacking at every step, Kos et al. \cite{kos2017delving} proposed to inject perturbations in every frame. Further, Lin et al. \cite{lin2017tactics} proposed to compute the preference of an agent taking the optimal action for better deploying perturbations. To better select the attack timing, several algorithms including Evolutionary Strategies\cite{yang2020enhanced}, Self-learning\cite{qiaoben2021strategically}, and Markov Decision Process\cite{russo2019optimal} have been developed. Apart from these, the transferability of adversarial perturbations across different DRL models is studied in \cite{behzadan2017vulnerability}. Benzadan et al. \cite{behzadan2017whatever} investigated the robustness and resilience of DRL models against adversarial attacks, and suggested that analysis of attack probability can facilitate secure design practices.

\section{Preliminaries}
\label{sec:pre}
\subsection{Problem Description}

There are two primary choices for air traffic control systems: centralized management and free flight. In the centralized system, the human controller is responsible for aircraft safety through tactical control actions. To keep safety, the aircraft of the system is always in a mixture of organized tracks and fixed airways. However, the continuous growth of air traffic demand has made centralized systems inefficient and more vulnerable to errors, and the free flight concept is a potential solution to those problems. Free flight means that the aircraft can freely plan its route between the defined entry point and exit point without referring to the fixed airway, and the required separation between aircraft is ensured by computer communication.

\begin{figure}[ht]
\centering
\includegraphics[width=0.5\linewidth]{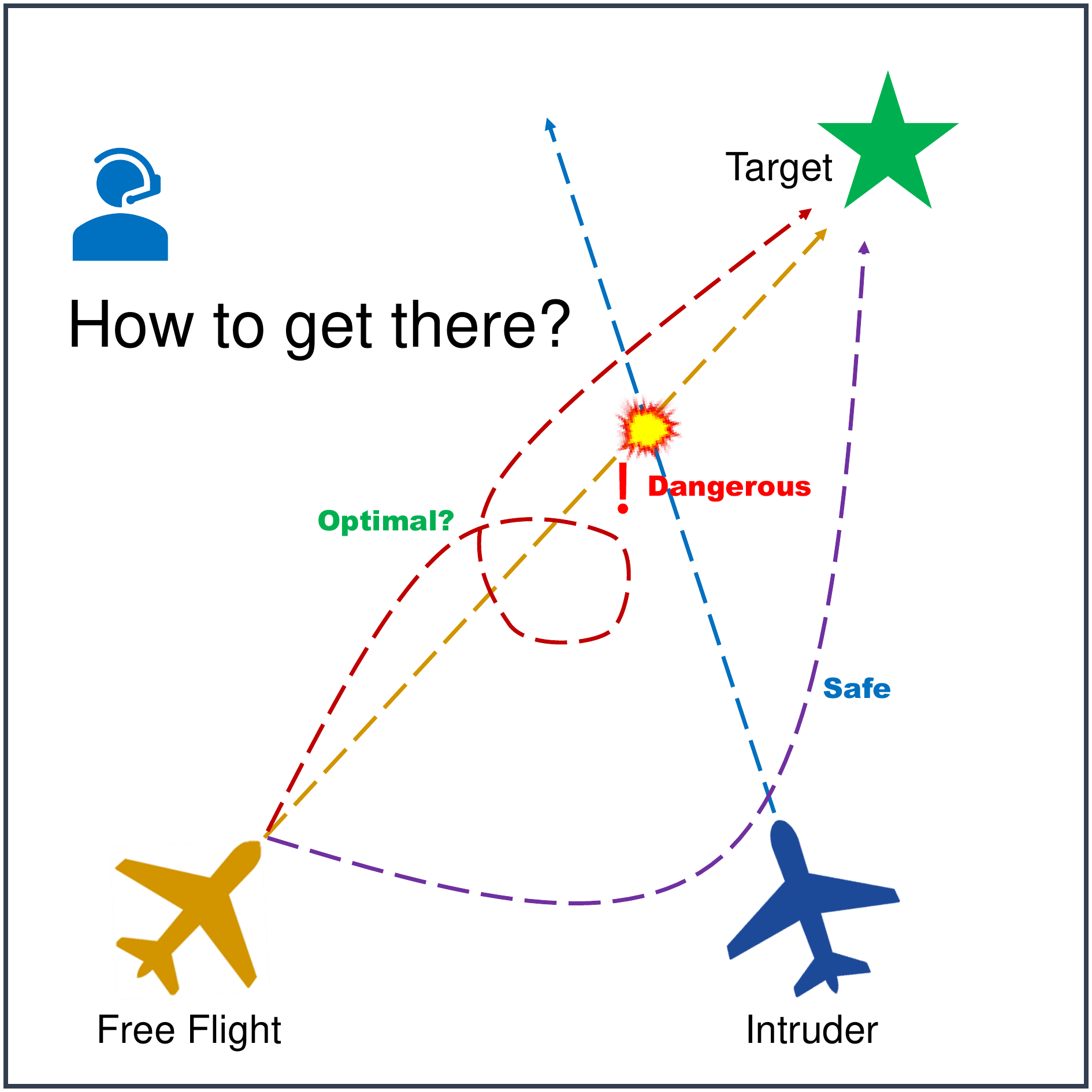}
\caption{The optimal control problem in this paper. A free flight is required to reach the destination as soon as possible. The yellow line is the fastest, but it will lead to potential conflicts. The purple line is safe and time-consuming. Our task is to find the optimal route marked as red color.}
\label{Fig: concept}
\end{figure}

In this work, we conduct a proof-of-concept study on the safety of DRL-based free flight control systems. To keep our problem as simple as possible, we use a DRL agent to control one free flight, and it is surrounded by several en-route flights controlled by the centralized system. Those en-route flights are moving on different airways. The task of the free flight is to reach the destination as soon as possible and keep the required separation between aircraft. We illustrate our control problem in Fig~\ref{Fig: concept}.

Our task can be mathematically represented as an optimal control problem as follows:
\begin{equation}
    \begin{split}
        \min_{u(t)} \quad \sum^T_{t=0} C&(\mathbf{x}_t,\mathbf{u}_t) \\
        s.t. \quad \dot{\mathbf{x_t}}  = F&(\mathbf{x_t}, \mathbf{u_t}, t), \\ g_i (\mathbf{x_t},   \mathbf{u_t}) \leq 0, & \quad i  = 1, 2, \cdots, I.
    \end{split}
    \label{eq:control_problem}
\end{equation}
where $C$ is the cost function (the time cost of the free flight), $g_i(\mathbf{x_t}, \mathbf{u_t}) \leq 0, i = 1, 2, \cdots, I$ are the constraints (minimum distance between the free flight and other controlled en-route flights) over state variables $\mathbf{x}$ (observations of the free flight) and action variables $\mathbf{u}$ (direction and acceleration).  

\subsection{Classic Deep Q Networks}

To solve sequential decision problems in Equation~\eqref{eq:control_problem}, we can learn estimates for the optimal value of each action, defined as the expected sum of future rewards when taking that action and following the optimal policy thereafter. Under a given policy $\pi$, the true value of an action $\mathbf{u}$ in a state $\mathbf{x}$ is
\begin{equation}
    Q^\pi(\mathbf{x}, \mathbf{u}) = \mathbb{E}_{ \mathbf{x}_{t} \sim E, \mathbf{u}_{t} \sim \pi}\left[\sum^\infty_{t=0} \gamma^k r_{t} \mid \mathbf{x}_0=\mathbf{x}, \mathbf{u}_0=\mathbf{u}\right]
\end{equation}
where $\gamma \in [0, 1]$ is a discount factor, $r_{t}$ is the reward at time point $t$. The optimal policy is then $\text{argmax}_\pi Q^\pi(\mathbf{x}, \mathbf{u})$. In DRL framework, the optimal policy $Q^\pi(\mathbf{x}, \mathbf{u})$ is parameterized by deep neural networks.

A deep Q network (DQN) is a deep neural network (DNN) that for a given state $x$ outputs a vector of action values $Q(\mathbf{x}, \cdot ; \theta)$, where $\theta$ are the parameters of the network. There are two key components of DQN. First, it uses a separate target network that is copied every $\tau$ steps from the regular network so that the target Q-values are more stable. Second, the agent adds all of its experiences to a replay buffer, which is then sampled uniformly to perform updates on the network. The learning objective for updating policy parameters $\theta^Q$ is to minimize the Bellman error:
\begin{equation}
\begin{split}
    L(\theta^Q) = \mathbb{E}_{\mathbf{x}_{t} \sim p^\beta, \mathbf{u}_{t} \sim \beta, r_t \sim E} & \left[ (Q(\mathbf{x}_t, \mathbf{u}_t ; \theta) - y_t)^2 \right], \\
    y_t = r(\mathbf{x}_{t}, \mathbf{u}_{t}) + & \gamma Q^{\text{target}}(\mathbf{x}_t, \mathbf{u}_t ; \theta),
    \end{split}
\label{eq:training}
\end{equation}
where $\beta$ is the arbitrary exploration policy, $\theta^Q$ parametrize the Q function, and replace $\theta$ every $\tau$ steps. 

\subsection{Adversarial Attacking on DRL}

DNNs are known to be vulnerable to adversarial example attacks. Furthermore, recent works show that adversarial attacks are effective when targeting DNN policies in the RL system. There are several ways to attack an RL agent: including adversarial perturbations on agents’ observations and actions, misspecification of the environment, adversarial disruptions on the agents and other adversarial agents.  This paper focuses on $l_\infty$-norm adversarial perturbations on agents’ observations $\mathbf{x}_t$ since this threat model is the most common approach to investigate the adversarial robustness of DRL. 

FGSM is one of the one-time attacks and is famous for its low time complexity. It is also the most common adversarial crafting algorithm in DRL. For an observation $\mathbf{x}$, $J(\mathbf{x}, \mathbf{u} ; \theta)$ denotes the cross-entropy loss between all actions $\mathbf{u}$ and the distribution that places all weight on the action with the highest Q value. $\bigtriangledown_\mathbf{x} J(\mathbf{x}, \mathbf{u} ; \theta)$ denotes the gradient of the loss function with respect to observation $\mathbf{x}$. For $l_\infty$-norm adversarial perturbations, the state $\mathbf{x}$ with optimal perturbation is calculated by
\begin{equation}
    \mathbf{x}^{\text{adv}} = \mathbf{x} + \epsilon \text{sign}\left(\bigtriangledown_\mathbf{x} J(\mathbf{x}, \mathbf{u} ; \theta)\right), 
\end{equation}
where $\mathbf{x}^{\text{adv}}$ is the adversarial observation, and $\epsilon$ is the constraint on the $l_\infty$-norm of perturbation. 

Except for the magnitude of the perturbation, another important attack dimension for the DRL system is the injection time. Since the spirit of adversarial attack is to apply a minimal perturbation to the observation to avoid detection. If the adversary perturbs the observation at every time during the control sequence, it is more likely to be detected. Therefore there have been numerous intends to reduce the number of times an adversarial example is crafted and used. Specifically, the adversary tries to minimize the expected cumulative reward $\mathbb{E} \left[\sum^{T -1}_{t=0} \gamma^t r_t \mid \mathbf{x}_0=\mathbf{x}, \mathbf{u}_0=\mathbf{u}\right]$, where the
attack strategy $k_t$ denotes whether at time step $t$ the perturbation should be applied ($k_t = 1$) or not ($k_t = 0$). Then the problem is: how can the adversary find an efficient attack strategy $[k_0, k_1, \cdots, k_{T-1}]$ that can minimize the expected cumulative reward and satisfy $\sum^{T -1}_{t=0} k_t \leq N$.

\section{Methodologies}
\label{sec:meh}
\subsection{Safety-aware Deep Q Networks}

The time-variant constraints $g_i (\mathbf{x}_t,\mathbf{u}_t) \leq 0$ on the action and the state in Equation~\eqref{eq:control_problem} are very important in real-world systems.
In many cases like ATCs, those constraints impose safety-critical refinements on control systems. The conventional DRL focuses on finding a policy that maximizes a long-term reward, which can be considered the negative value of the cost function $C$ in Equation~\eqref{eq:control_problem}. Therefore, DRL will likely violate the constraints during the learning process since it focuses on maximizing the long-term reward. This characteristic is problematic for any RL algorithm deployed in the real 
ATC system, as violating the constraints can lead to unsafe behaviors. The common way to handle this problem is to redefine the reward function by adding a penalty term to the cost function:
\begin{equation}
    r(\mathbf{x}_t,\mathbf{u}_t) = w_c C(\mathbf{x}_t,\mathbf{u}_t) + w_gG(\mathbf{x}_t,\mathbf{u}_t),
    \label{eq:reward_engineer}
\end{equation}
where $w_c<0$ and $w_g>0$ are reward weights, $G(\mathbf{x}_t,\mathbf{u}_t)$ is a reward function inferred from the constraints $g_i (\mathbf{x}_t,  \mathbf{u}_t) \leq 0$ in Equation~\eqref{eq:control_problem}. The process for finding desirable $w_c$, $w_g$, and $G(\mathbf{x}_t,\mathbf{u}_t)$ is handled by the so-called ``reward engineering'' process, which is usually very tedious and requires expert knowledge. 

Even after the ``reward engineering'' process, DRL agents pursuing efficiency and optimality of the cumulative engineered reward may also execute unsafe actions. The reason is that they are agnostic with respect to safety measure $G(\mathbf{x}_t,\mathbf{u}_t)$ in Equation~\eqref{eq:reward_engineer}. Equation~\eqref{eq:reward_engineer} is unidentifiable in the sense that given $r$
we cannot recover $C$ and $G$ uniquely. In some cases, an action with a large reward return $\sum^\infty_{k=0} \gamma^k R(\mathbf{x}_{t+k+1},\mathbf{u}_{t+k+1})$ may lead to states with very small safety measure $G(\mathbf{x}_t,\mathbf{u}_t)$, and agents that are
agnostic with respect to $G(\mathbf{x}_t,\mathbf{u}_t)$ may execute unsafe actions without deliberateness.

\begin{figure}[ht]
\centering
\includegraphics[width=0.49\textwidth]{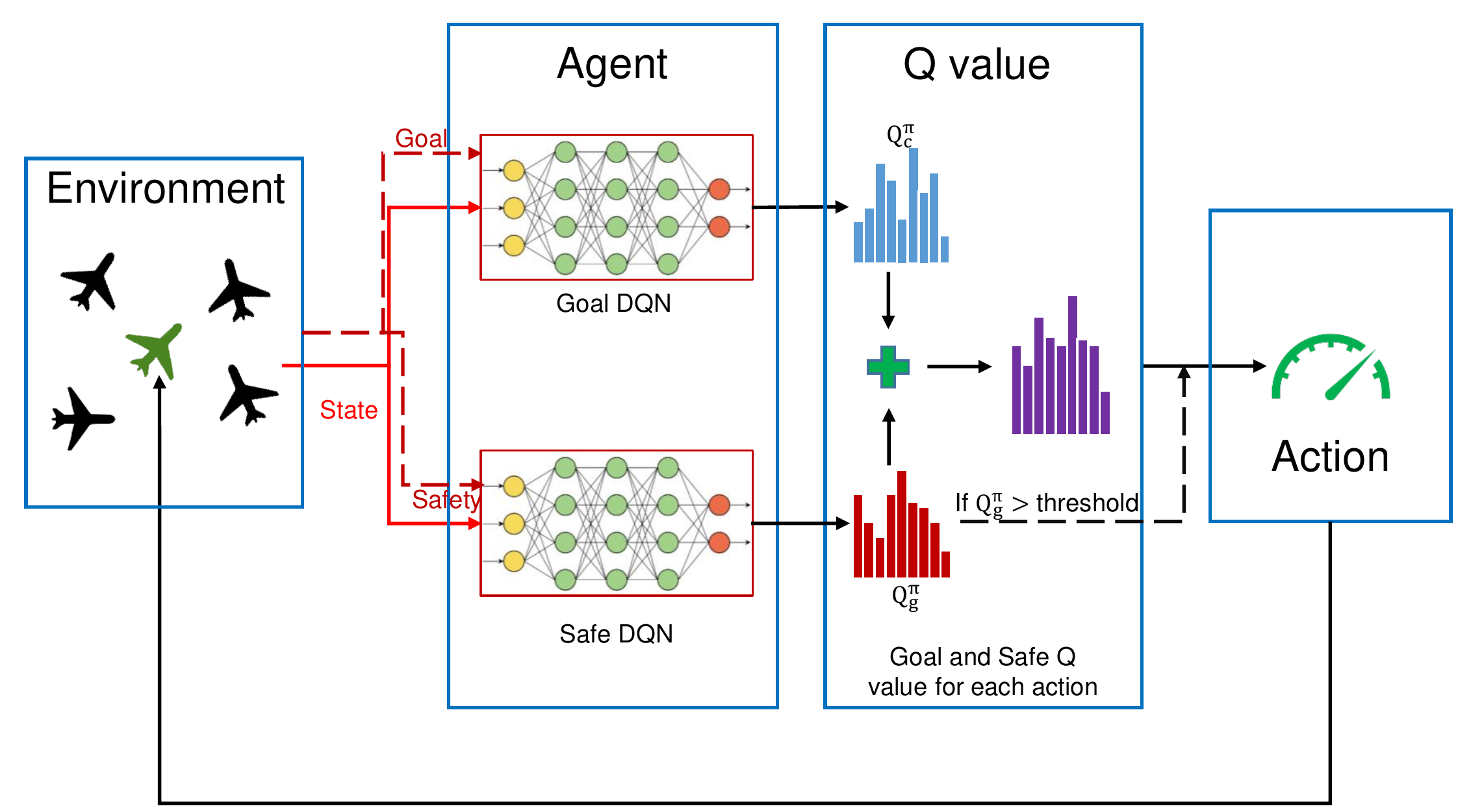}
\caption{Illustration of our safety-aware DQN structure. The structure consists of two deep neural networks, which receive the same state. The goal DQN uses the goal reward $r^c$ to estimate the goal Q values $Q^c_\pi$, while the safe DQN uses safety reward $r^g$ to estimate the safety Q values $Q^g_\pi$. We pick up the actions with safety Q values larger than a threshold, and the one with the highest overall Q value is executed. }
\label{Fig: structure}
\end{figure}

We adopt a safety-aware deep Q-learning (SafeDQN) model to overcome the difficulties mentioned above. We simply assume that the observed reward is generated from a weighted sum with true cost $C$ and constrained function $G$. The idea of our model is to decompose the overall Q values estimation into the sum of primary Q values and a safety Q values. We have
\begin{equation}
\begin{split}
    Q^\pi(\mathbf{x}, \mathbf{u}) &= \mathbb{E}_{ \mathbf{x}_{t} \sim E, \mathbf{u}_{t} \sim \pi}\left[\sum^\infty_{t=0} \gamma^k r_{t} \mid \mathbf{x}_0=\mathbf{x}, \mathbf{u}_0=\mathbf{u}\right] \\
    &= \mathbb{E}_{ \mathbf{x}_{t} \sim E, \mathbf{u}_{t} \sim \pi}\left[\sum^\infty_{t=0} \gamma^k (w_c C_{t} + w_g G_{t}) \mid \mathbf{x}_0=\mathbf{x}, \mathbf{u}_0=\mathbf{u}\right].
    \end{split}
\end{equation}
 Without loss of generality, let $r^c_t = w_c C_t$, $Q^\pi_c =\mathbb{E}_{ \mathbf{x}_{t} \sim E, \mathbf{u}_{t} \sim \pi}\left[\sum^\infty_{t=0} \gamma^k r^c_t \mid \mathbf{x}_0=\mathbf{x}, \mathbf{u}_0=\mathbf{u}\right] $, $r^g_t = w_g G_t$ and $Q^\pi_g =  \mathbb{E}_{ \mathbf{x}_{t} \sim E, \mathbf{u}_{t} \sim \pi}\left[\sum^\infty_{t=0} \gamma^k r^g_t \mid \mathbf{x}_0=\mathbf{x}, \mathbf{u}_0=\mathbf{u}\right] $. Given $E(X+Y) = E(X) + E(Y)$ and $E(wX) = wE(X)$, we have
\begin{equation}
    Q^\pi =  Q^\pi_c +  Q^\pi_g.
\label{eq: Q_sum}
\end{equation}
Drawing insights from Equation~\eqref{eq: Q_sum}, we use two deep neural networks to estimate primary (goal) Q-value $ Q^\pi_c$ and safety Q-value $ Q^\pi_g$, respectively. By estimating the safety Q-value in the proposed framework, we can learn about and verify safety by evaluating the safety value $Q^\pi_g$ of the current state. In conventional DQN, the action is selected by $\mathbf{u}_t = \text{argmax}_u Q^\pi(\mathbf{x_t}, \mathbf{u})$. For our safety-aware DQN, we can select action by $\mathbf{u}_t = \text{argmax}_u \left(Q^\pi_c(\mathbf{x_t}, \mathbf{u}) + Q^\pi_g(\mathbf{x_t}, \mathbf{u})\right)$ if the safety value $Q^\pi_g(\mathbf{x_t}, \mathbf{u_t})$ is higher than a threshold. Otherwise, the agent will choose another safer action even if the overall Q value of the action is the highest. The model illustration is in Figure~\ref{Fig: structure}.

Another positive outcome of the formulation is that the original coupled Q-value learning problem is decomposed into two separate Q-value learning problems, making our formulation more amenable to safety or goal-oriented attacks. For example, we can generate state perturbations according to the safety Q-value $Q^\pi_g(\mathbf{x_t}, \mathbf{u_t})$:
\begin{equation}
    \mathbf{x}^{\text{adv}} = \mathbf{x} + \epsilon \text{sign}\left(\bigtriangledown_\mathbf{x} J^g(\mathbf{x}, \mathbf{u} ; \theta^g)\right), 
    \label{eq:att_safe}
\end{equation}
where $J^g(\mathbf{x}, \mathbf{u} ; \theta)$ denotes the cross-entropy loss between all actions $\mathbf{u}$ and the distribution that places all weight on the action with highest $Q^\pi_g(\mathbf{x_t}, \mathbf{u})$ value, $\theta^g$ are the parameters of the safety Q networks. In our experiments, we find that this safety-oriented attack can easily lead to aircraft conflicts. 

\subsection{State Representation for Free Flight}
\label{sec: state_rep}

In DRL framework, the state, at any given time, captures the information needed by
the policy in order to understand the current condition of the environment and decide what action to take. Despite learning in an end-to-end manner, tho choice of state representation can still affect the problem difficulty and contribute to the brittle nature of DRL performances \cite{andrychowicz2021matters}. 

\begin{figure}[ht]
\centering
\includegraphics[width=0.38\textwidth]{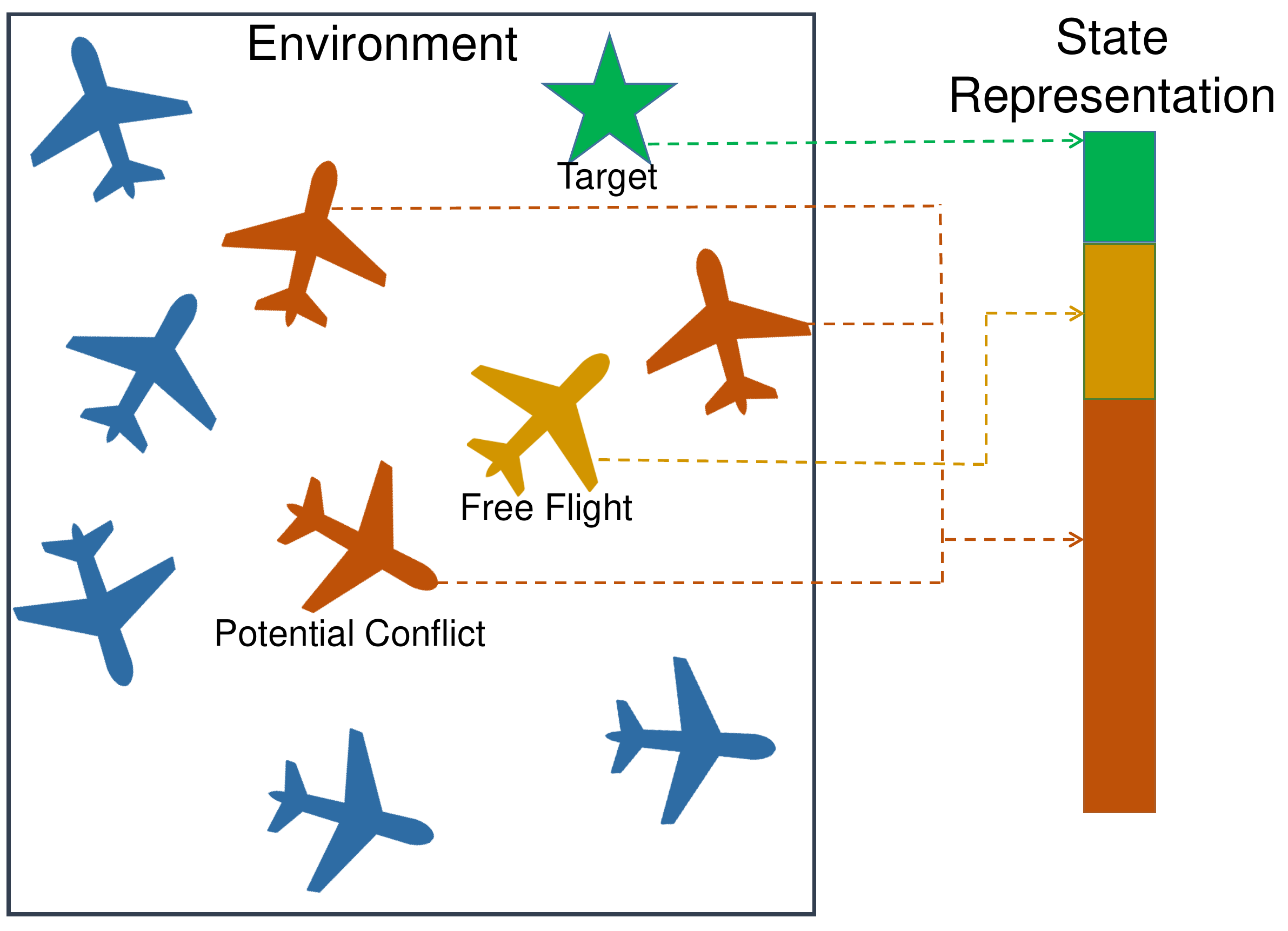}
\caption{The state representation in this work. In this example, we only use the aircraft's three nearest neighbors as state observations. The aircraft far away from the free flight are not considered. }
\label{Fig: state}
\end{figure}

In this work, we are designing state observation of the DRL agent for the free flight. Traditionally, centralized ATC systems monitor the location, speed, and direction of all aircraft in their assigned airspace by radar and issue instructions that pilots are required to obey. The restricted nature of the centralized ATC system aids controllers in projecting potential conflicts. Aircraft usually follow established routes. These restrictions do not work for the free flight. We argue that monitoring all aircraft information for free flight is unnecessary. To make a decision for free flight, the agent only needs to consider the information of surrounding aircraft having the potential to generate conflicts. Moreover, we also find that the redundant information brought by aircraft far away from the controlled free flight makes that the DRL agent hard to learn a desirable policy. In particular, the state for the free flight is represented as
\begin{equation}
    \mathbf{x} = \left\{ \{ \mathbf{s}_i : i\in  \mathit{N}(o) \}, \mathbf{s}_o, \mathbf{t}  \right\},
    \label{eq:state}
\end{equation}
where $\mathit{N}(o)$ denotes the set of $k$ free flight's nearest neighbor with $\| \mathit{N}(o) \| = k$, $\mathbf{s}_i$ is the state of its surrounding aircraft, $\mathbf{s}_o$ is the state of the controlled free flight, and $\mathbf{t}$ is the location of the target. By this means, the free flight only needs to monitor its $k$ nearest neighbors at each time point, eliminating the information redundancy brought by aircraft without conflict threats. The state representation is illustrated in Figure~\ref{Fig: state}. Another benefit is that this state representation is invariant to the aircraft density and volume of the airspace, making the trained DRL agent can adapt to environments with variable densities and surrounding aircraft. We will show those benefits in our experiments. 

\begin{algorithm}[htp]
 \caption{Training algorithm for the proposed SafeDQN-X}
 \begin{algorithmic}[1]
 \renewcommand{\algorithmicrequire}{\textbf{Input:}}
 \renewcommand{\algorithmicensure}{\textbf{Initialize:}}
 \REQUIRE Training episodes $I$, Nearest intruder number $k$, learning rate decay $d$, Update every $\tau$ step
 
 \ENSURE  $Q_c$, ${Q_c}^{target}$, $Q_g$, ${Q_g}^{target}$, Environment, Replay Buffer $D$ 
  \FOR {$i = 1$ to $I$}
  \STATE Initialize the free flight environment with $X$ air routes, and $X$ is a random number
  \WHILE{episode not done}
  \STATE Obtain the $k$ nearest aircraft, get current state $\mathbf{x_t}$ according to Equation~\eqref{eq:state}\\
  \STATE 
  Calculate Q values, $Q(\mathbf{x_t}, \mathbf{u}) = Q_c(\mathbf{x_t}, \mathbf{u}) + Q_g(\mathbf{x_t}, \mathbf{u})$\\
  \STATE Get the undetermined optimal action, $u_t^* = \arg\max_{\mathbf{u}} Q(\mathbf{x_t}, \mathbf{u})$  \\
  \WHILE{$Q_g(\mathbf{x_t}, u_t^*) < \delta$}
 \STATE $\mathbf{u} = \mathbf{u}\setminus u_t^*$ \\
 ${u_t}^* = \arg\max_{\mathbf{u}}Q(\mathbf{x_t}, \mathbf{u})$\\
  \ENDWHILE
\STATE With probability $\eta$ select a random action ${u_t}^*$ \\
  \STATE Perform $u_t^*$ and receive $r_{g, t}$, $r_{c, t}$  and $\mathbf{x_{t+1}}$\\
  Store transitions $(\mathbf{x_{t}}, \mathbf{x_{t+1}}, u_t^*, r_{g, t}, r_{c, t}, done)$ in $D$\\
  \STATE $\mathbf{x_{t}} = \mathbf{x_{t+1}}$
  \STATE Sample random minibatch of transitions $(\mathbf{x_{j}}, \mathbf{x_{j+1}}, u_j^*, r_{g, j}, r_{c, j}, done)$ from $D$\\
  \STATE Update the weights of $Q_c$ and $Q_g$ based on the gradients of Equation~\eqref{eq:training}\\
   \IF{$t \mod  \tau == 0$}
  \STATE $Q_c^{target} = Q_c$, $Q_g^{target} = Q_g  $
  \ENDIF
  \ENDWHILE
  \ENDFOR
 \end{algorithmic}
 \label{algo:SafeDQN}
\end{algorithm}

Given the safety-awareness and state representation designs, we conclude the training algorithm for SafeDQN in Algorithm~\ref{algo:SafeDQN}. We train SafeDQN on environments with random $X$ air routes, thus we name the SafeDQN trained by Algorithm~\ref{algo:SafeDQN} as SafeDQN-X. In Algorithm~\ref{algo:SafeDQN}, the safety threshold value $\delta$ controls how our agent chooses the safe action. However, finding a proper $\delta$ value is not easy. In particular, the safety awareness of our agent changes (the value of $Q_g$) during the training process. Therefore, to ensure that our agent can output an action, we set $\delta$ as the mean value of $Q_g(\mathbf{x_t}, \mathbf{u})$. During training, the action is selected by an $\epsilon$-greedy policy that follows the greedy and safety policy with probability $1 - \eta$ and selects a random action with probability $\eta$, and $\eta$ decays during training process. 

\subsection{Attack Timing}
\label{sec: att_time}
It has been shown that analysis of adversarial attack probability can facilitate safe design practices for real-world control systems. Given that safety is the priority for ATCs, we also study the performance of our SafeDQN under adversarial attacks. Specifically, we select Strategically-Timed (ST) Attack method \cite{lin2017tactics} to attack our model because it is highly efficient and explainable. The method defines a relative action preference value $p$ to choose attack timing. The value $p$ computes the preference of the agent in taking the most preferred action over the least preferred action at the current state $\mathbf{x_t}$. The degree of preference to action depends on the trained DQN policy. A large ${p}$ value implies that the agent strongly prefers one action over the other. We briefly describe the relative action preference function ${p}(\cdot)$ for attacking the agents trained by the DQN algorithms below. 

When an agent strongly prefers a specific action
(The action has a relatively high probability), in words, it is critical to perform the action; otherwise, the accumulated reward will be reduced. Based on this intuition, the preference $p$ is computed by:
\begin{equation}
    p\left(\mathbf{x_t}\right)=\max _{\mathbf{u}} \frac{e^{\frac{Q\left(\mathbf{x_t}, \mathbf{u}\right)}{T}}}{\sum_{k} e^{\frac{Q\left(\mathbf{x_t}, u({k})\right)}{T}}}-\min _{\mathbf{u}} \frac{e^{\frac{Q\left(\mathbf{x_t}, \mathbf{u}\right)}{T}}}{\sum_k e^{\frac{Q\left(\mathbf{x_t}, u({k})\right)}{T}}},
\label{eq:att_time}
\end{equation}
where the value of temperature ${T}$ is set to ${1}$ in our experiments. Equation~\eqref{eq:att_time} converts the computed Q-values of actions into a probability distribution over actions using the softmax function with the temperature constant ${T}$. In the ST attack, the adversary attacks the DQN agent at time step ${t}$ when the relative action preference ${p}$ has a value greater than a threshold parameter ${\beta}$.

\begin{figure}[ht]
\centering
\includegraphics[width=0.35\textwidth]{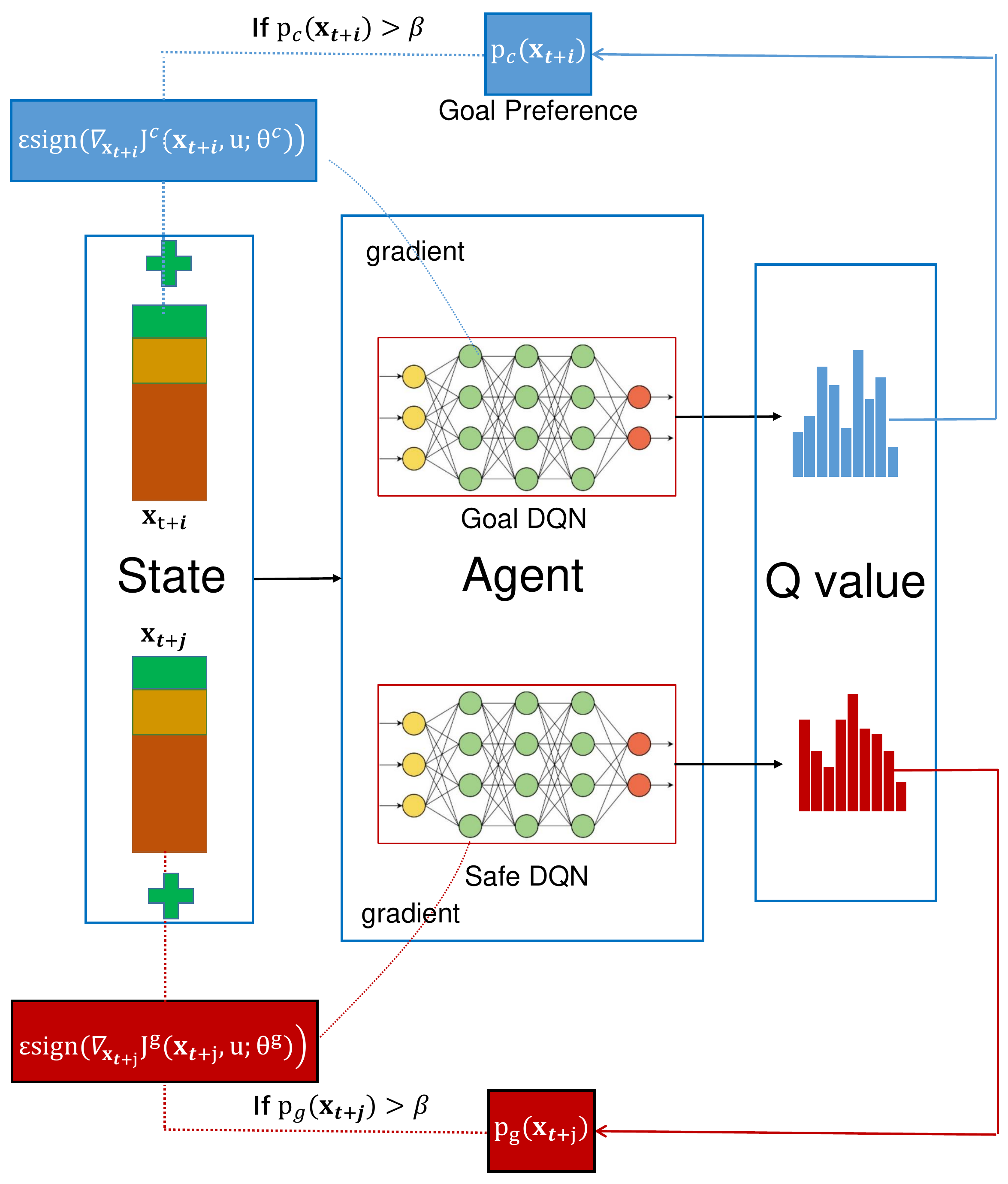}
\caption{The ST framework for safety-oriented and goal-oriented attacks. For each state, we compute the safety and goal preferences. If one of them is greater than a threshold $\beta$, the adversarial will impose a perturbation on the state. The perturbation is calculated by the corresponding Q networks. }
\label{Fig: att}
\end{figure}

Furthermore, rather than attacking the overall Q values, we conduct goal and safety-oriented attacks on the proposed SafeDQN. Our SafeDQN produces goal Q values $Q_c$ and safe Q values $Q_g$. If we want to conduct the safety-oriented attack, we first compute the safety preference $p_g(\mathbf{x_t})$ using $Q_g$ based on Equation~\eqref{eq:att_time}. If $p_g(\mathbf{x_t}) > \beta$, the state perturbation $\mathbf{x_t}^\text{adv}$ is then produced by Equation~\eqref{eq:att_safe}, $\mathbf{x_t}^\text{adv}$ is then fed into the agent to mislead its action. The goal-oriented attack is vice versa. We summarize the ST framework for safety-oriented and goal-oriented attacks in Figure~\ref{Fig: att}.

\section{Experiments}
\label{sec:exp}
In this section, we evaluate the performance of our proposed model, SafeDQN-X, for free flight control with extensive experiments and compare the results with conventional DRL methods using several metrics. First, we modified the simulation environment developed by Yang et al.\cite{yang2018autonomous} for training all DRL models. The primary goal of the modified environment is to control a free flight through a series of actions so that the aircraft can quickly arrive at the destination while avoiding potential conflict with other intruder aircraft. Then, after training all DRL agents in our environment, we present a quantitative and qualitative evaluation of the adversarial attack on the trained agent. Finally, we explain the behavior of SafeDQN-X on free flight control.

\subsection{Experiments Setup}

\subsubsection{Environment}

\begin{figure}[ht]
\centering
\subfigure[Initial state]{
\label{Fig:env.sub.1}
\includegraphics[width=0.23\textwidth]{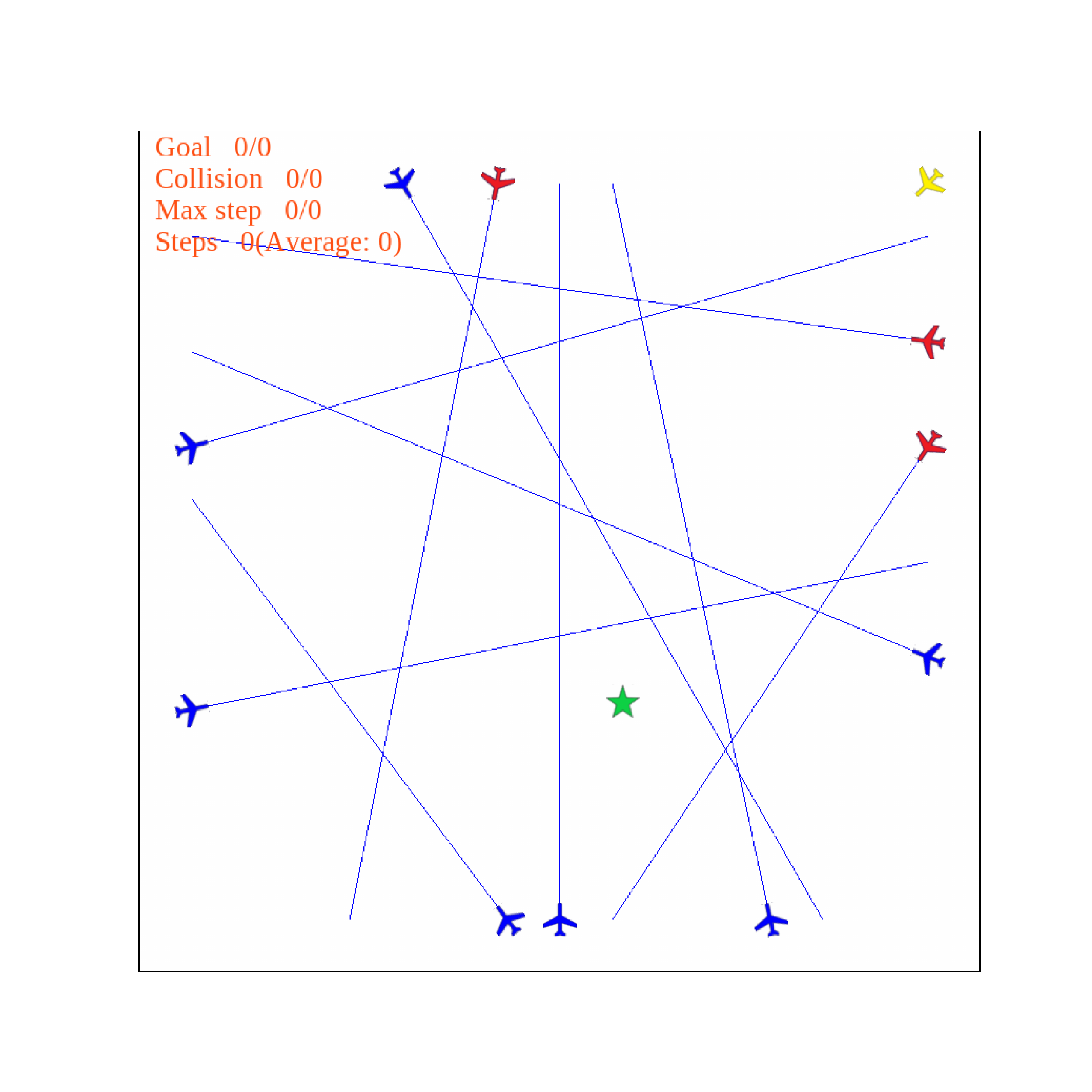}}
\subfigure[Reach goal]{
\label{Fig:env.sub.2}
\includegraphics[width=0.23\textwidth]{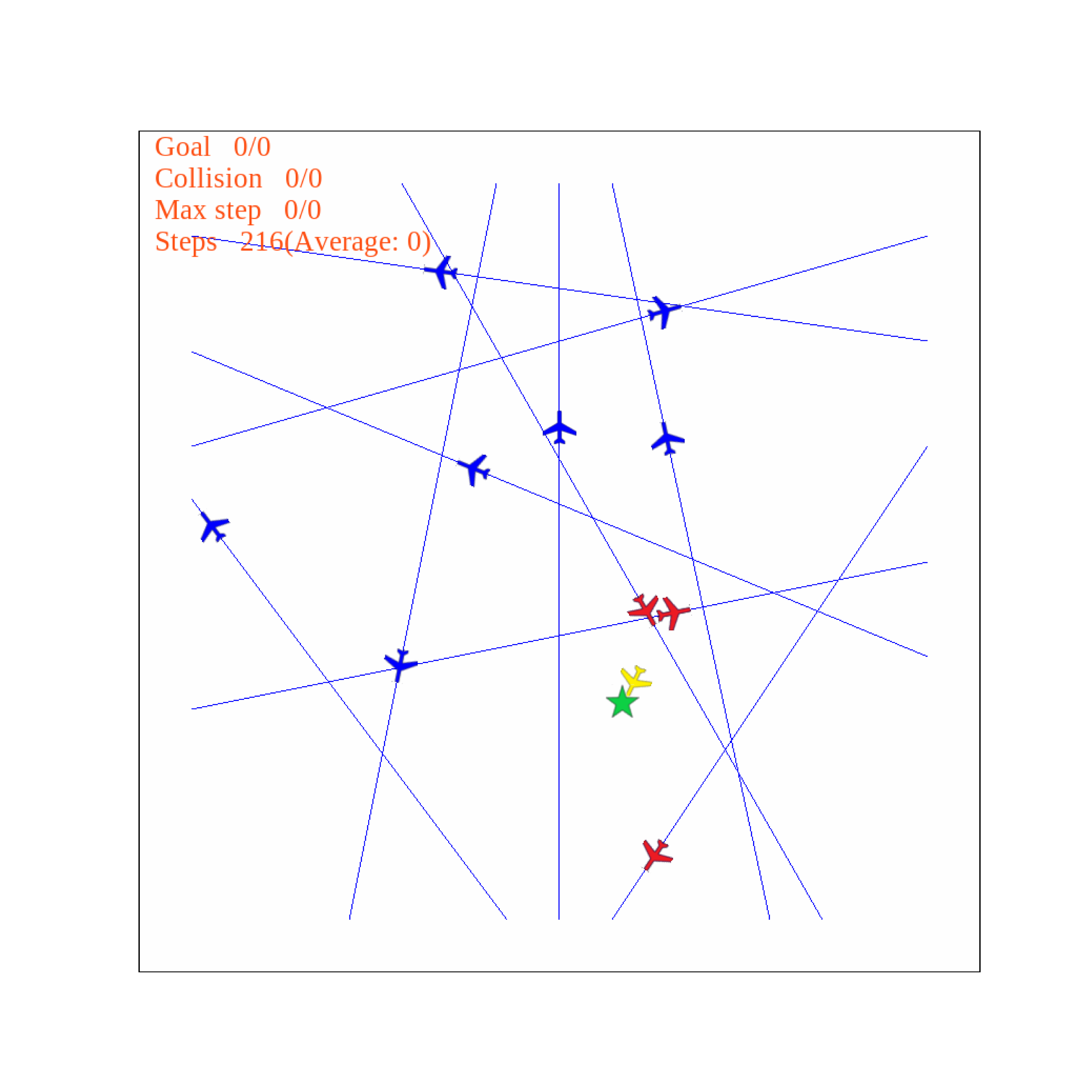}}
\caption{Experimental environment. (a) Initialize a random goal point and random numbers of fixed routes, with one plane on each route. The free flight appears from one of the four corners of the environment. (b) The optimal terminal state: the free flight successfully reaches the final goal.}
\label{Fig:env}
\end{figure}

The experimental environment is set as shown in Figure~\ref{Fig:env}. There are random numbers ($3\sim 25$) of fixed routes. Each route has an en-route aircraft (red) with a fixed heading and speed. We assume that central controllers control all aircraft on those routes. An aircraft (the yellow one) controlled by the DRL agent (DQN) randomly appears from four corners of the environment with an initial fixed heading angle. It acts a free flight, whose task is to reach the random target (green star) as soon as possible, as shown in figure~\ref{Fig:env.sub.2}, and avoid disturbing others. Since this is a proof-of-concept study, we keep the environment as simple as possible. All planes are operated in a 2D environment, and our agent only needs to control the free flight's speed and direction.

{\textbf{State:}}As stated in Section~\ref{sec: state_rep},  Each state observed by the agent includes the position and speed of its $k$ nearest neighbors, its own position, speed and heading direction information, and the position information of the target point, which can be concluded as
\begin{equation}
        \mathbf{x}=\left\{\{x_i,y_i,v_{x_i},v_{y_i}: i \in \mathit{N}(o) \},x_o,y_o,v_{x_o},v_{y_o},s,h,x_g,y_g\right\}
\end{equation}
where $\mathit{N}(o)$ denotes the set of $k$ free flight's nearest neighbor with $\| \mathit{N}(o) \| = k$, $\{x_i,y_i,v_{x_i},v_{y_i} \}$ represents the position and speed of each intruder, $s$ and $h$ represent speed and heading direction of the free flight, respectively, $(x_g,y_g)$ denotes the target position. Thus we have a state vector $\mathbf{x} \in \mathrm{R}^{4k + 8}$. 

{\textbf{Action:}}  The action space of the agent is limited to both heading direction and speed changes, and the agent can either accelerate/decelerate, turn to the left/right, or maintain the current track and speed. Thus, the action space $\mathbf{u}$ is modeled as a set of 9 discrete actions, which can be divided into two parts:
\begin{equation}
\begin{split}
        \mathbf{u}=  \mathbf{A_s} & \times  \mathbf{A_h} \\
        \mathbf{A_s}=\{acc,& keep, dec\},  \\
        \mathbf{A_h}=\{left,& keep, right\}. 
\end{split}
\end{equation}
where $\mathbf{A_s}$ controls the speed, and $\mathbf{A_h}$ controls the heading direction. 

{\textbf{Reward:}} The reward function needed to be designed to reflect the goal of this task:
\begin{itemize}
\item Safe separation;
\item Minimizing the delay in arriving at the final metered position;
\item Choosing the optimal route.
\end{itemize}
The reward is set as follows:
\begin{equation}
    r =
\begin{cases}
1 & \text{if the controlled aircraft reaches the target,}\\
-1& \text{if a collision is happened,}\\
-0.5 & \text{if a conflict is happened,}\\
-0.0001 & \text{otherwise (step penalty).}
\end{cases}
\end{equation}
As stated before, we decompose the reward into a primary reward and a safety reward. Here, the primary reward $r^c$ is
\begin{equation}
    r^c =
\begin{cases}
1 & \text{if the controlled aircraft reaches the target,}\\
-0.0001 & \text{otherwise (step penalty).}
\end{cases}
\end{equation}
and safety reward $r^g$ is
\begin{equation}
    r^g =
\begin{cases}
-1& \text{if a collision is happened,}\\
-0.5 & \text{if a conflict is happened,}
\end{cases}
\end{equation}

{\textbf{Terminal State:}} Termination in the episode could be achieved in three different ways:

\begin{itemize}
  \item Goal reached (optimal): The controlled aircraft reached its final target position, maintained safe separation, and avoided the collision.
  
  \item Max step reached (feasible): The aircraft did not arrive at their final metered position in 1000 steps.
  
  \item Collision: The controlled flight collided with another flight.
\end{itemize}

\subsubsection{DQN Structure}

With respect to our model architecture, we adapt two Dueling DQN structures \cite{wang2016dueling} to estimate the primary Q values $Q^\pi_c$ and the safety ones $Q^\pi_g$, respectively. The output layer is split into two streams: one for estimating the state-value $V(\textbf{x})$ and the other for estimating state-dependent action advantages. After the two streams, the last module of the network combines the state-value and advantage output $A(\textbf{x}, \textbf{u})$. Then those two values are aggregated by 
\begin{equation}
Q(\mathbf{x}, \mathbf{u}) = V(\mathbf{x}) + \left(A(\mathbf{x}, \mathbf{u}) - \max_{u' \in \| A \|} A(\mathbf{x}, \mathbf{u}) \right),
\label{eq:value_duel}
\end{equation}
where $V(\mathbf{x})$ is possible to show which states are valuable. This is also beneficial for explaining the behaviors of our agent. For example, the safety value $V^g(\mathbf{x})$ can be considered as the situational safety awareness of the current state. A state $\mathbf{x}$ with an extremely low safety value would be considered dangerous. For the hyperparameters, our agent receives 3 nearest aircraft's states, thus the goal and safe DQNs have an input layer with dimension $3\times 4 + 8 = 20$, three hidden residual layers with dimension 128, 64 and 32, respectively, an output value with dimension 1, and an output advantage with dimension 9. We use Relu function as the activation for all the hidden layers. We train for 20000 episodes using Adam optimizer with learning rate 0.0001, and learning rate dacay 0.9999.

\subsubsection{Baselines}

To show the benefits of safe-awareness design and $k$-nearest neighbor state representation, we compare SafeDQN-X with the following models.
\begin{itemize}
  \item \textbf{DQN}: The traditional DQN model with a coupled Q value is trained on environments with a fixed number of routes (using all surrounding intruders' state as inputs). 
  
  \item \textbf{SafeDQN}: The safety-aware DQN model is trained on environments with a fixed number of routes.
  
  \item \textbf{DQN-X}: The DQN model with $k$-nearest neighbor state representation is trained on environments with a random number of routes. 
\end{itemize}

The evaluation metrics we choose include safety-oriented metrics, goal-oriented metrics and the {\textbf{average score}} (average total reward). Specifically, the safety-oriented metrics include the {\textbf{collision rate}} (Near Mid Air Collision, NMAC) and {\textbf{conflict rate}} (loss of separation). The goal-oriented metrics include the {\textbf{max-step reached}} (the free-flight intruder fail to reach the target in 1000 steps) and the {\textbf{average step}} to reach the target.

\subsection{Performance without adversarial attack}

This section describes the experimental results of SafeDQN-X under normal cases in which no adversarial attack is imposed. We randomly generate 500 cases with five fixed airways and 500 cases with ten fixed airways and evaluate the metrics of all DQNs using the same scenarios. The aim of the experiments is to answer the following two questions:
\begin{itemize}
    \item \textbf{$\mathbf{Q1:}$ Is the safety-aware design really aware of the collision risks?}
    \item \textbf{$\mathbf{Q2:}$ Can local state representation improve our model?}
\end{itemize}

\begin{figure}[ht]
\centering
\includegraphics[width=\linewidth]{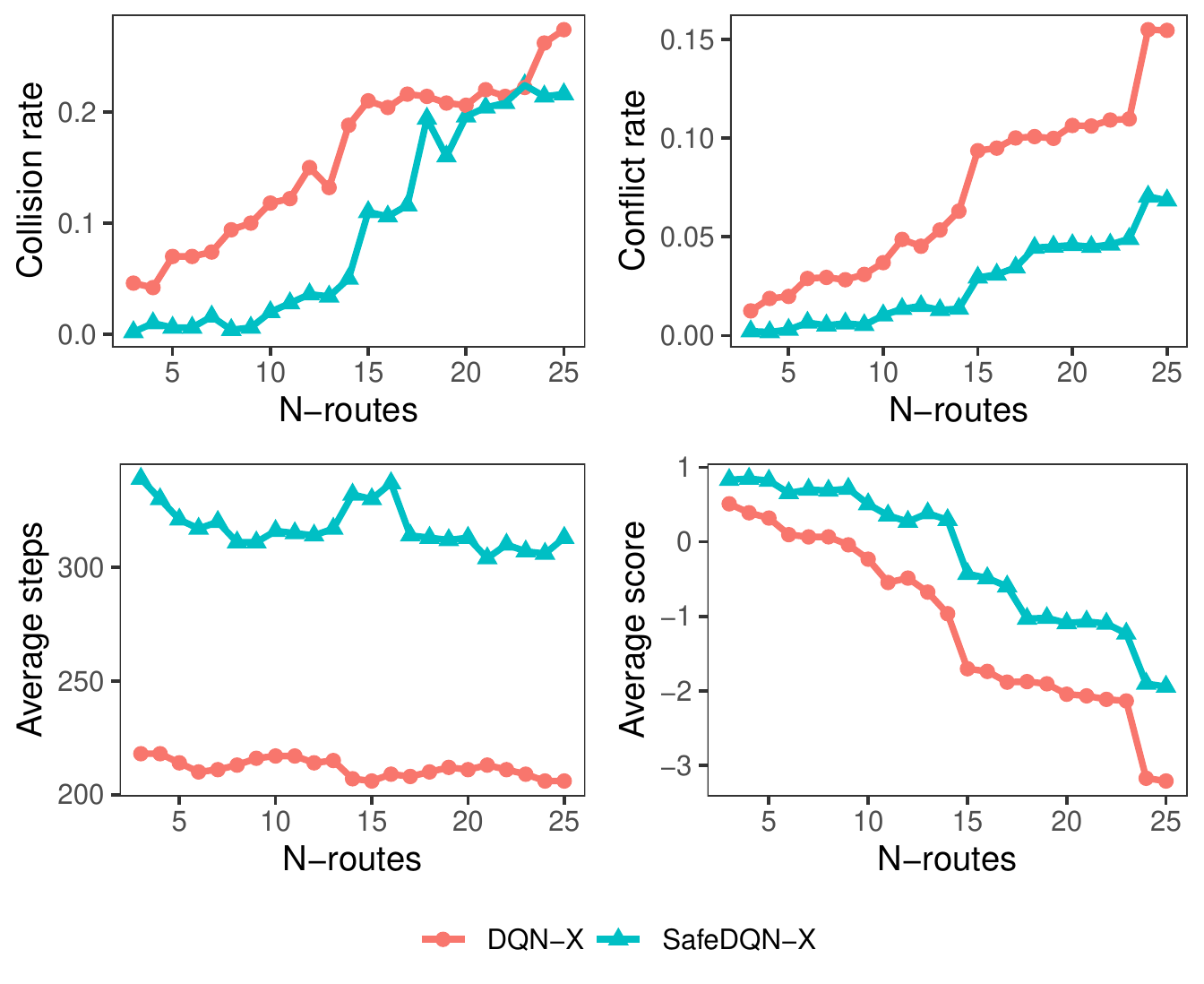}
\caption{Performance comparison of DQN-X and SafeDQN-X with respect to number of surrounding routes. The collision rate and conflict rate of SafeDQN-X are much lower than that of DQN-X. However, the average steps to reach the goal of SafeDQN-X are higher, illustrating that SafeDQN-X's decision-making is more conservative.}
\label{fig.4}
\end{figure}

Table~\ref{tab.table.1} summarizes the results of all DQN models under 5 and 10 routes scenarios. The results produced by our SafeDQN-X are clearly promising. For example, SafeDQN-X achieved 2.0\% collision rate under ten route scenarios. However, the density is relatively high in this case, making this scenario very challenging. All other counterparts' collision rates are above 10\% under this scenario, suggesting that DRL models without local state representation and safe-awareness design could not make a safe decision under high-density airspace. Furthermore, we can find that regardless of the fixed route training or random route training, the collision rate and conflict rate of the agents controlled by SafeDQN are much lower than those controlled by traditional DQN. However, their probability of reaching the maximum steps and the average steps to reach the goal is higher, indicating that safety-awareness design makes our model more conservative to potential risks, resulting in relatively lower efficiency. Based on the above results, SafeDQN can easily get a higher average score, as we give high weights on collision in reward design. The sacrifice of efficiency is worthy, as collision and conflict would cause incredible damage to air mobility. In addition, we compare the performance of DQN-X and SafeDQN-X on cases with route numbers ranging from 3 to 25 in Figure~\ref{fig.4}, and the same conclusion can be reached. Therefore, we can get the answer to $\mathbf{Q1}$: SafeDQN is indeed more risk conscious. An interesting finding in Figure~\ref{fig.4} is that the advantage of SafeDQN over DQN in terms of conflict rate decreased when the route numbers were above 20. The airspace is overcrowded under those cases with more than 20 routes, making the CR tasks challenging to any agents. 

\begin{table*}[!ht]
\centering
\caption{Four methods' Performance comparisons on 5 and 10 routes scenarios}
\resizebox{\textwidth}{!}
{
\begin{tabular}{c c c c c c}
\bottomrule
& collision rate  & conflict rate & max-step reached & average steps & average score\\
\hline
\hline
DQN-5 & 5.2\%  & 1.69\% & ${0.6\%}$ & $\mathbf{207}$ & 0.43\\

SafeDQN-5 & ${3.0\%}$ & ${0.56\%}$ & 2.8\%  & 316 & ${0.65}$\\

DQN-X5 & 7.0\%  & 1.98\% & $\mathbf{0.2\%}$ & ${214}$ & 0.32\\

SafeDQN-X5 & $\mathbf{0.6\%}$  & $\mathbf{0.29\%}$ & 2.6\% & 321 & $\mathbf{0.82}$\\
\midrule

DQN-10 & 15.0\%  & 5.52\% & $\mathbf{0.0\%}$ & $\mathbf{203}$ & ${-0.63}$\\

SafeDQN-10 & ${11.8\%}$  & ${1.71\%}$ & 3.8\% & 325 & ${-0.02}$\\

DQN-X10 & 11.8\%  & 3.69\% & $\mathbf{0.0\%}$ & ${217}$ & ${-0.23}$\\

SafeDQN-X10 & $\mathbf{2.0\%}$  & $\mathbf{1.02\%}$ & 1.8\% & 316 & $\mathbf{0.50}$\\

\toprule
\end{tabular}
}
\label{tab.table.1}
\end{table*}

\begin{figure}[ht]
\centering
\includegraphics[width=0.7\linewidth]{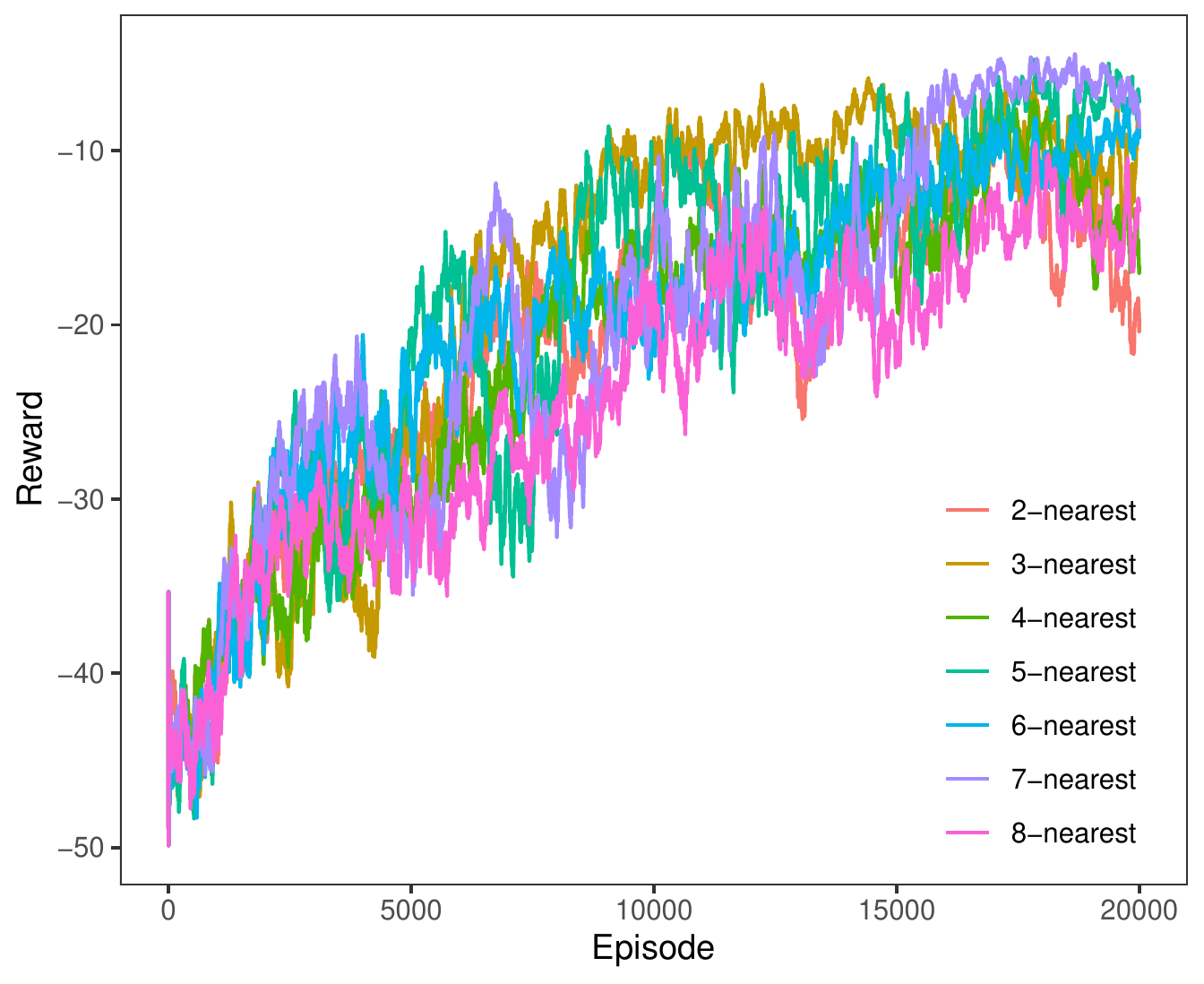}
\caption{Training curve of different SafeDQN-X models with different number of nearest neighbors (Evaluated on 10-route scenarios).}
\label{k-nearest-training-curve}
\end{figure}

As for $\mathbf{Q2}$, we can also conclude from Table~\ref{tab.table.1} that SafeDQN-X achieves better performance under the same scenarios compared with SafeDQN. In addition, DQN-X is also better than DQN. The local state representation makes our model adapt to scalable environments with variable numbers of surrounding intruders. It also requires less information, which in turn, reduces the parameters of DQN models. We mark here that the answer for $\mathbf{Q2}$ is quite interesting. The primary aim of local-state representation is to filter out redundant information. The results resonate with swarm intelligence in real-world, e.g., the real birds' information about the world is limited to their nearby flockmates. At the same time, their synchronized group behaviors are both beautiful to watch and intriguing to contemplate. Therefore the local sensing design has been incorporated in most multi-agent systems since the earliest ``boids'' system\cite{reynolds1987flocks}. Our results suggest that making the free flight agent omniscient is unnecessary. Even worse, the redundant information brought by intruders without conflict potential is harmful to our DQN agent. In addition to numerical metrics, we also report the learning curves of SafeDQN-X models with different numbers of nearest neighbors in Figure~\ref{k-nearest-training-curve}. We find that the models using three neighbors converge faster than those with more neighbors.

\begin{figure}[ht]
\centering
\includegraphics[width=\linewidth]{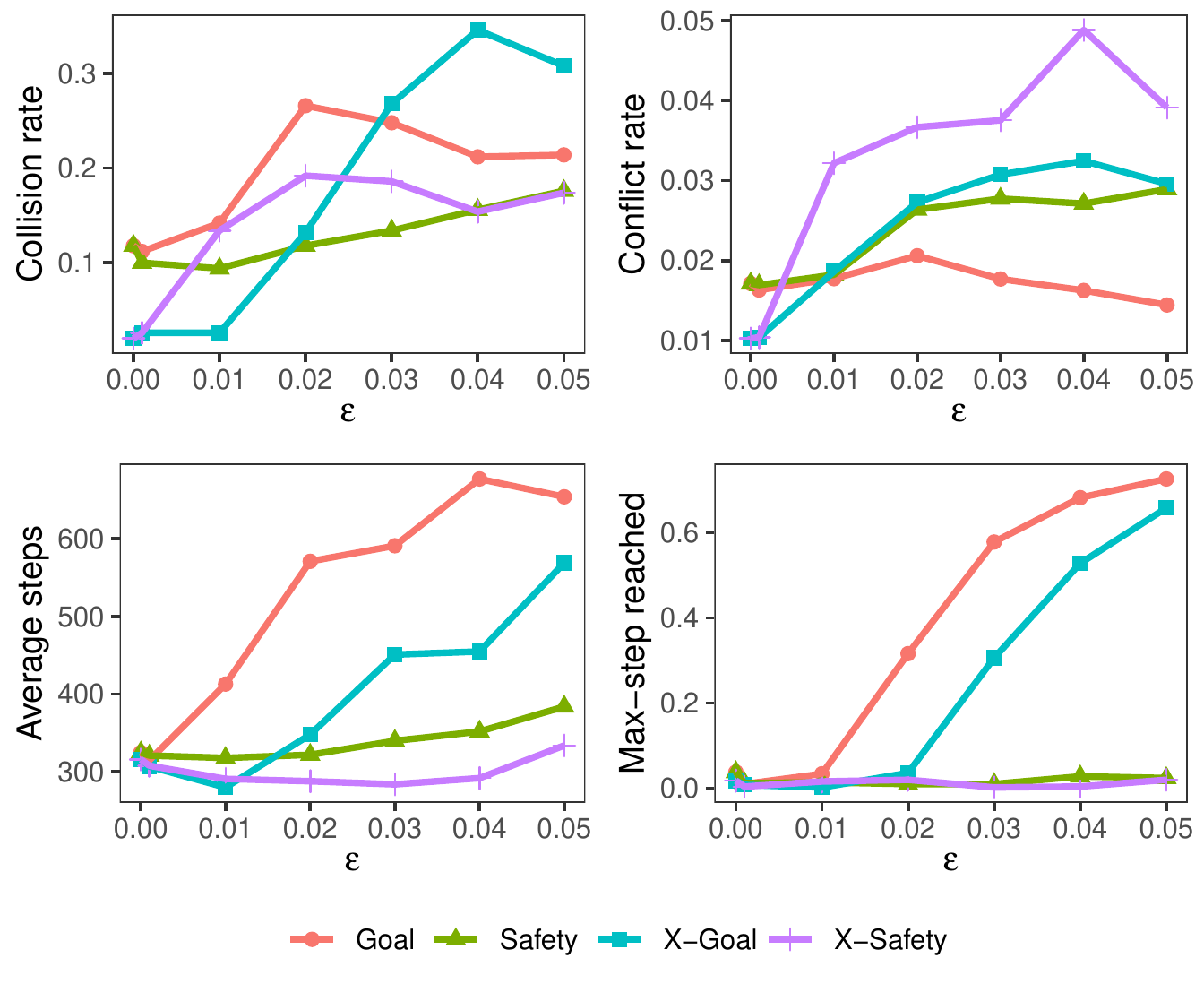}
\caption{Safety-oriented and goal-oriented Uniform attacks with frequency equal to 1 on SafeDQN and SafeDQN-X. Goal and Safety are for SafeDQN, and X-Goal and X-Safety are for SafeDQN-X. }
\label{Uniattack_diff_epsilon}
\end{figure}

\begin{figure}[ht]
\centering
\includegraphics[width=\linewidth]{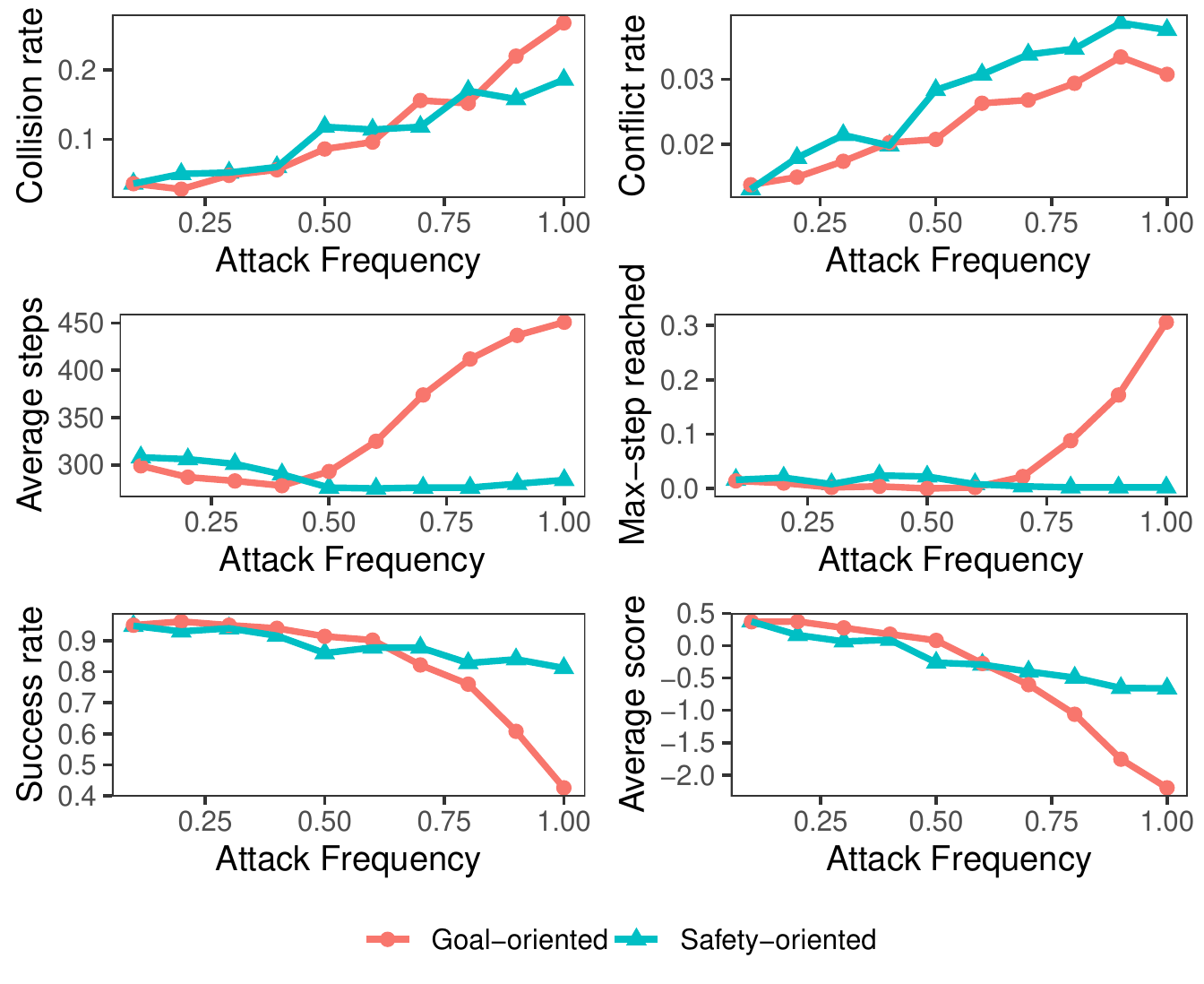}
\caption{Safety-oriented and goal-oriented Uniform attacks with different frequency with magnitude $\epsilon = 0.03$.}
\label{Uniattack2type}
\end{figure}

\subsection{Performance with adversarial attack}

Here, we show the results of our proposed algorithm under 10 routes scenarios with adversarial attacks. We use safety-oriented and goal-oriented adversarial attack methods proposed in Section~\ref{sec: att_time} to attack our agents. The aim of this experiment is twofold: First, we aim to explore the performance of DRL-based ATC agents under adversarial attacks. Second, we want to verify whether the proposed attack scheme can act on the chosen specific target. Since we separate the reward signal into a safe and goal one, the experiment also provides further evidence about the proposed reward separation design.

We first analyze how safety and goal-oriented metrics vary with the magnitude of attack perturbation. Figure~\ref{Uniattack_diff_epsilon} shows the trend of \textbf{collision rate}, \textbf{conflict rate}, \textbf{average step} and \textbf{max-step reached} for different values of magnitude ($\epsilon$ in Equation~\eqref{eq:att_safe}). Here we impose adversarial attacks at every time step (Uniform attack) on SafeDQN and SafeDQN-X, without using the proposed attack timing strategy given in Equation~\eqref{eq:att_time}. We notice that the magnitude of the goal-oriented attack can significantly increase \textbf{average step} and \textbf{max-step reached}. In contrast, the safety-oriented attack almost does not affect those two metrics. Such an outcome was expected. However, it is particularly interesting that goal-oriented attacks can also significantly increase both \textbf{collision rate} and \textbf{conflict rate}. This should not come as a surprise since the goal-oriented attack can significantly increase the flying time of the controlled intruder. The increased conflicts and collisions are a by-product of the increased flying time. Combining this with the experimental results, we conclude that the proposed attack scheme can solely act on the chosen target.

\begin{figure*}[ht]
\centering
\subfigure[Safety-oriented]{
\label{STvsUni.Safe}
\includegraphics[width=0.46\linewidth]{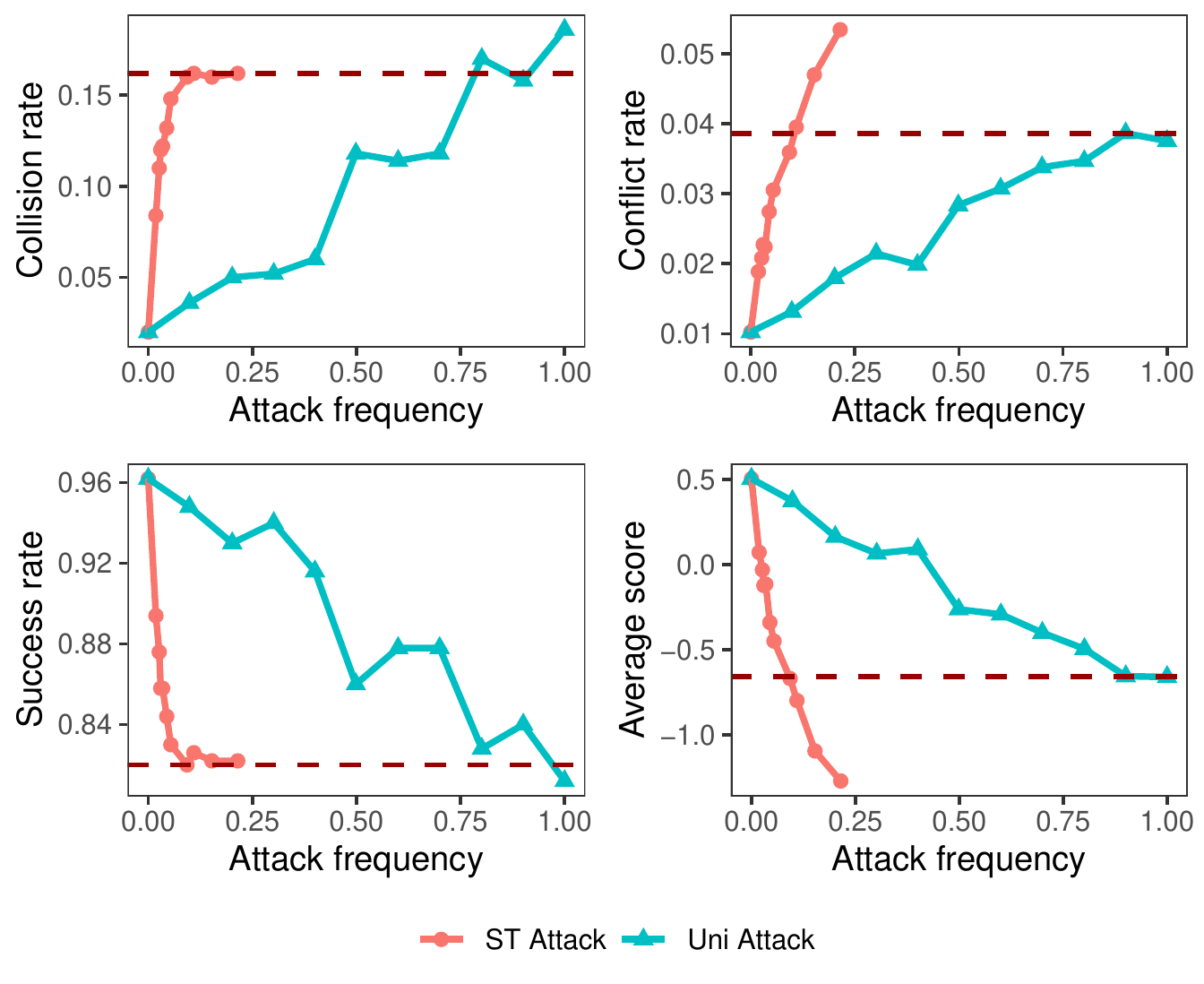}}
\subfigure[Goal-oriented]{
\label{STvsUni.Goal}
\includegraphics[width=0.46\linewidth]{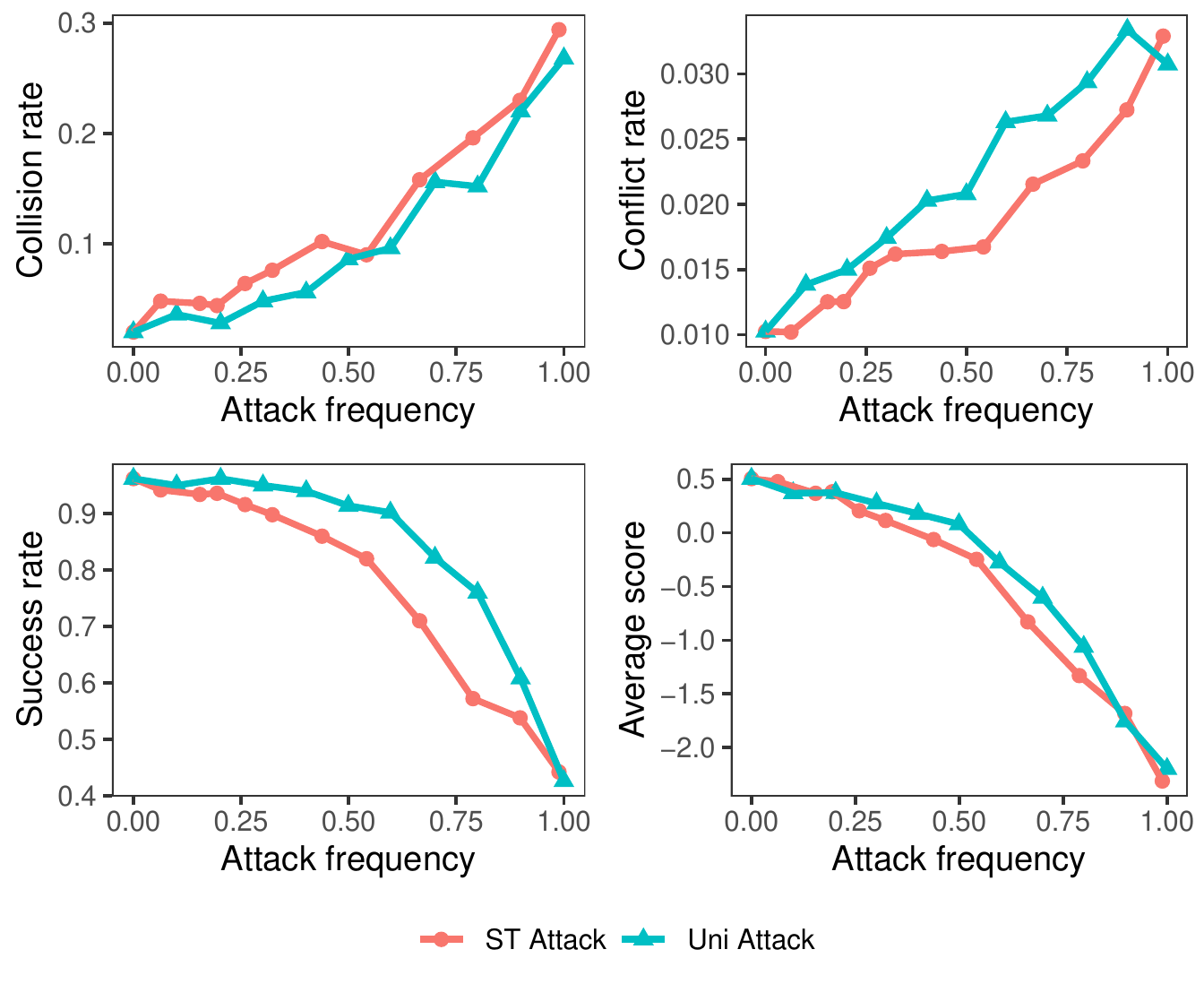}}
\caption{Performance comparison of Uniform Random Attack and the proposed Strategically-Timed Attack. (a) Safety-oriented. The performance of ST Attack is better than that of Uni Attack. (b) Goal-oriented. The performance of ST Attack is the same as Uni Attack.}
\label{STvsUni}
\end{figure*}

We further analyze how different metrics vary with the attack frequency ($\frac{Attack Time}{Overall Time}$). We fix perturbation magnitude $\epsilon$ to 0.03 and randomly choose the attack time. In Figure~\ref{Uniattack2type}, we immediately see that the \textbf{average step} and \textbf{max-step reached} significantly increased when goal-oriented attack frequency is higher than 0.5. In contrast, the frequency of safety-oriented attacks has no impact on the goal-related metrics. The frequencies of safety-oriented and goal-oriented attacks can both increase conflicts and collisions. Those observations are concordant with the behavior of Figure~\ref{Uniattack_diff_epsilon}. We also find that the random attack time strategy is ineffective in time dimension, e.g., attacking 50\% steps is required to increase the \textbf{average step}. 

Next, we examine the efficiencies of the proposed Strategically-Timed (ST) attack method in Section~\ref{sec: att_time}. It should be noted that the value of $\beta$ mainly controls the attack frequency of ST attacks. When $\beta$ decreased to a certain value, the attack frequencies stopped increasing. We compare the proposed ST attack with the Uniform random attack under different frequencies in Figure~\ref{STvsUni}. The magnitude of the perturbation is set to 0.03. As shown in the Figure, the proposed ST attack method can reach the same effect as the 100\% uniform attack timing at desirable attacked timing rates (25\% for collision rate, 10\% for the other three metrics). The results verify the effectiveness of the proposed attack approach.



\begin{figure}[ht]
\centering
\includegraphics[width=\linewidth]{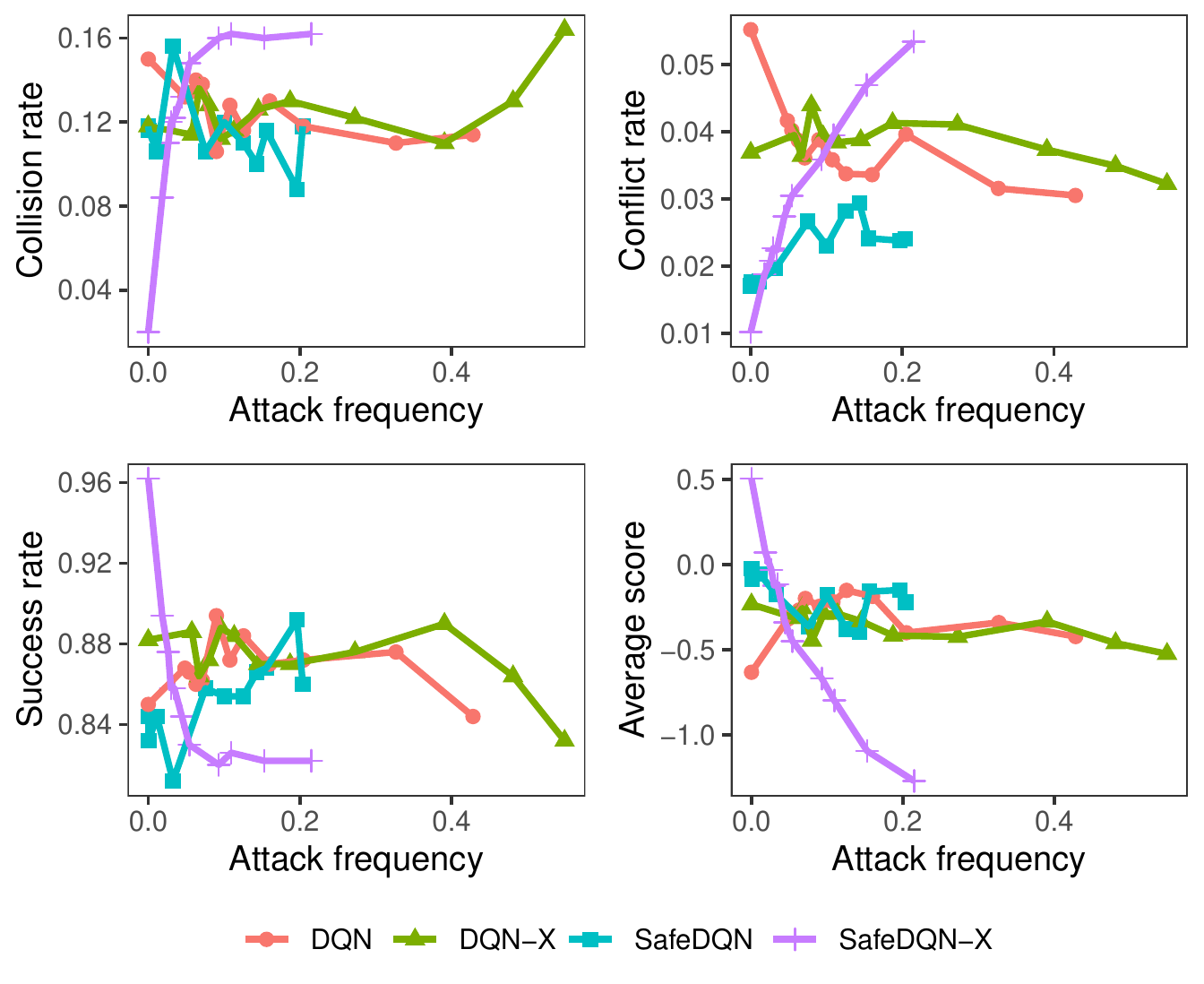}
\caption{ST attacks on four different models. We conduct safety-oriented attacks on SafeDQN and SafeDQN-X and overall attacks on DQN and DQN-X.}
\label{STAttack.SafeDQN-X}
\end{figure}

We further compare the effects of ST attack on different models, and the experimental results are shown in Figure~\ref{STAttack.SafeDQN-X}. An interesting finding is that only the metrics of SafeDQN-X (the purple curves) significantly changed with respect to the increased attack frequencies. For DQN and DQN-X, it is reasonable since the goal and safety attacks are not decoupled. At the same time, the ST attack is also inefficient for SafeDQN. Recall that the attack timing of the ST attack is determined by the output Q-values of DQN models. The SafeDQN model did not give satisfactory metrics in Table~\ref{tab.table.1} when there is no attack, indicating that it fails to estimate the accurate safety Q-values. In this case, a more powerful model is more vulnerable under adversarial attacks. The smartest agent has done the dumbest thing, which provides an accurate Q-values for achieving highly efficient attacks. We suggest that training a powerful learning-based controller for real-world applications like ATCs is not the end of the research. The controller's security under adversarial attacks also requires intensive attention. 

\subsection{SafeDQN-X behavior visualization}

\begin{figure*}[ht]
\centering

\subfigure[A safe state]{
\label{Fig.safe.state}
\includegraphics[width=0.23\linewidth]{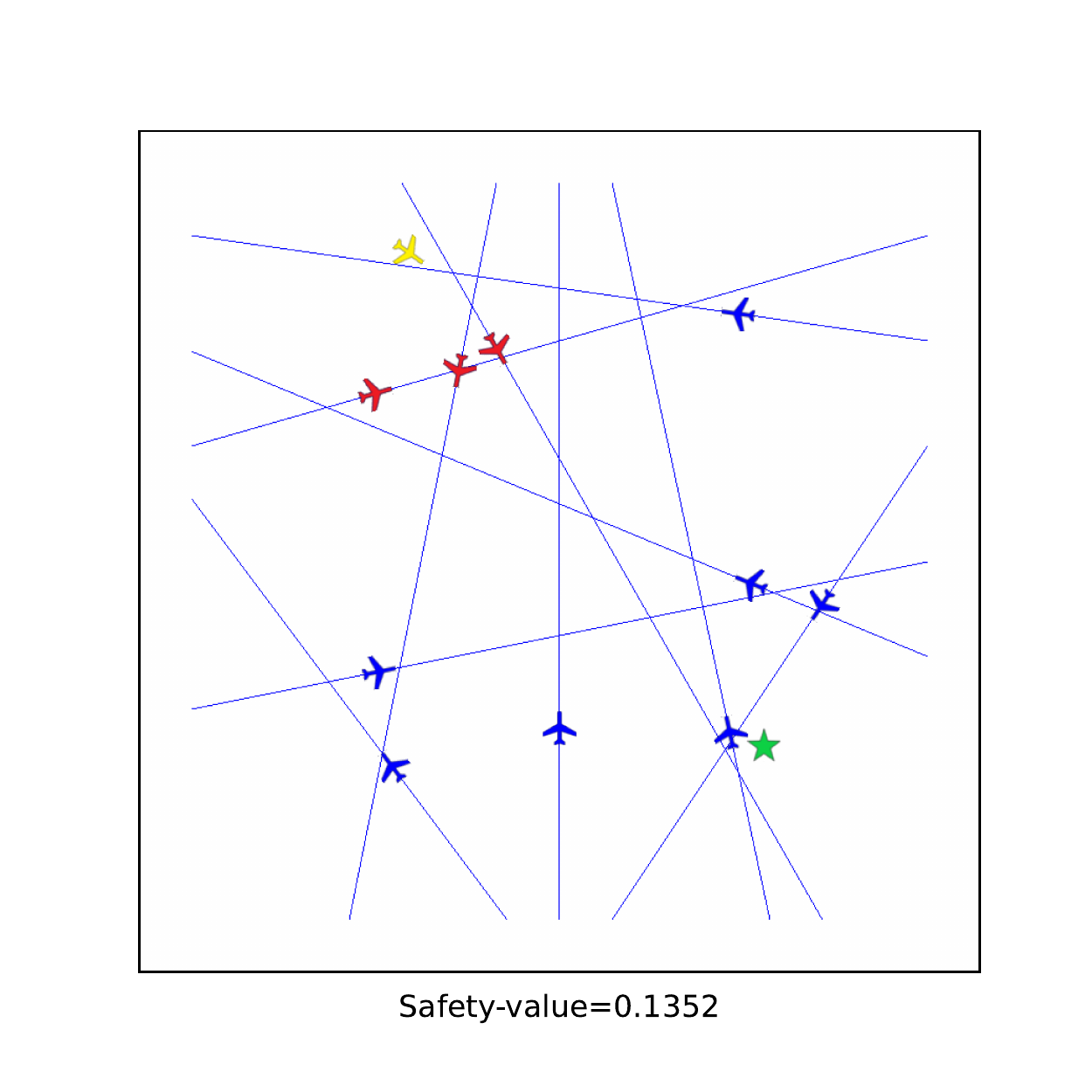}}
\subfigure[The overall action probability map under the safe state]{
\label{Fig.safe.heatmap.SafeDQN-X}
\includegraphics[width=0.23\linewidth]{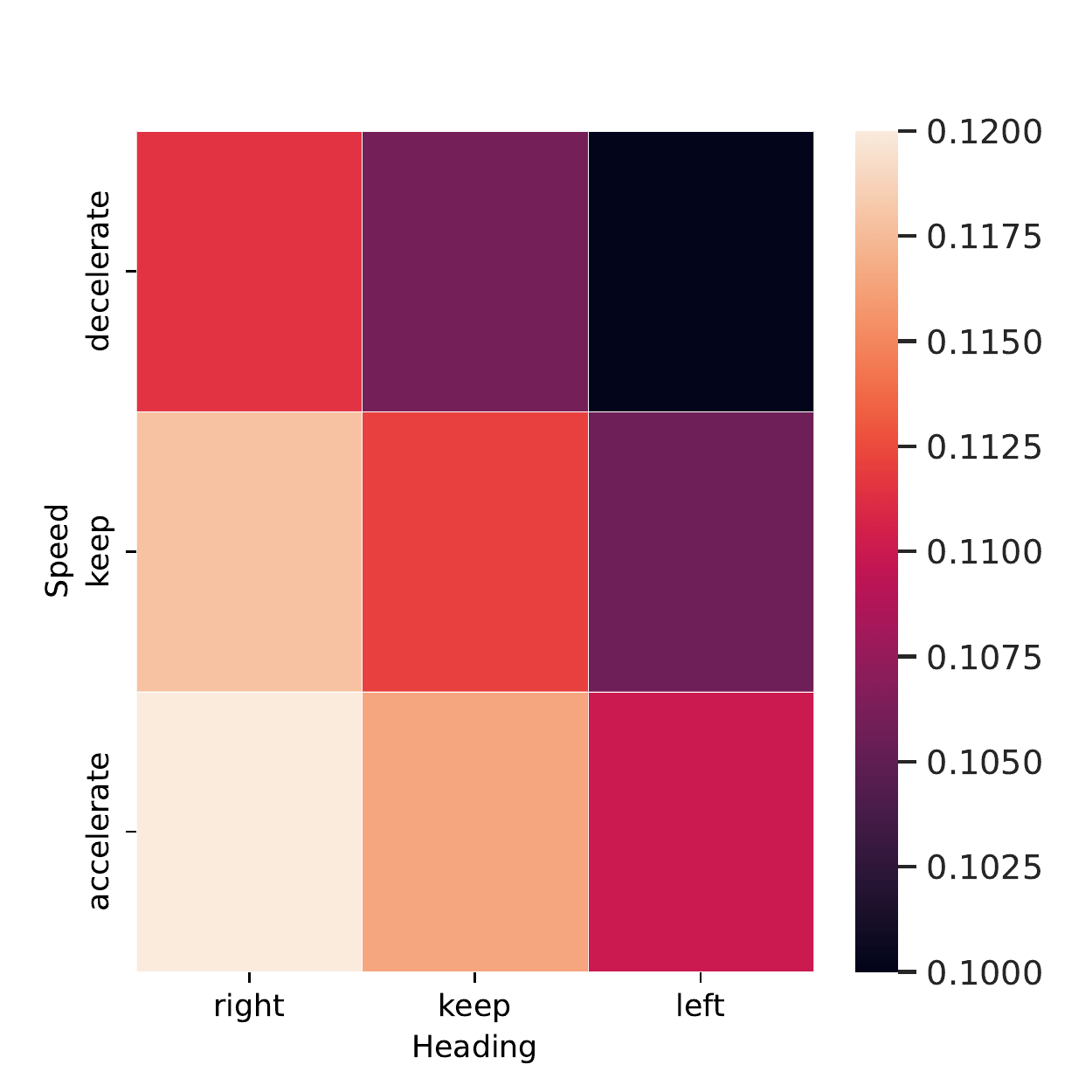}}
\subfigure[The goal action probability map under the safe state]{
\label{Fig.safe.heatmap.GoalDQN}
\includegraphics[width=0.23\linewidth]{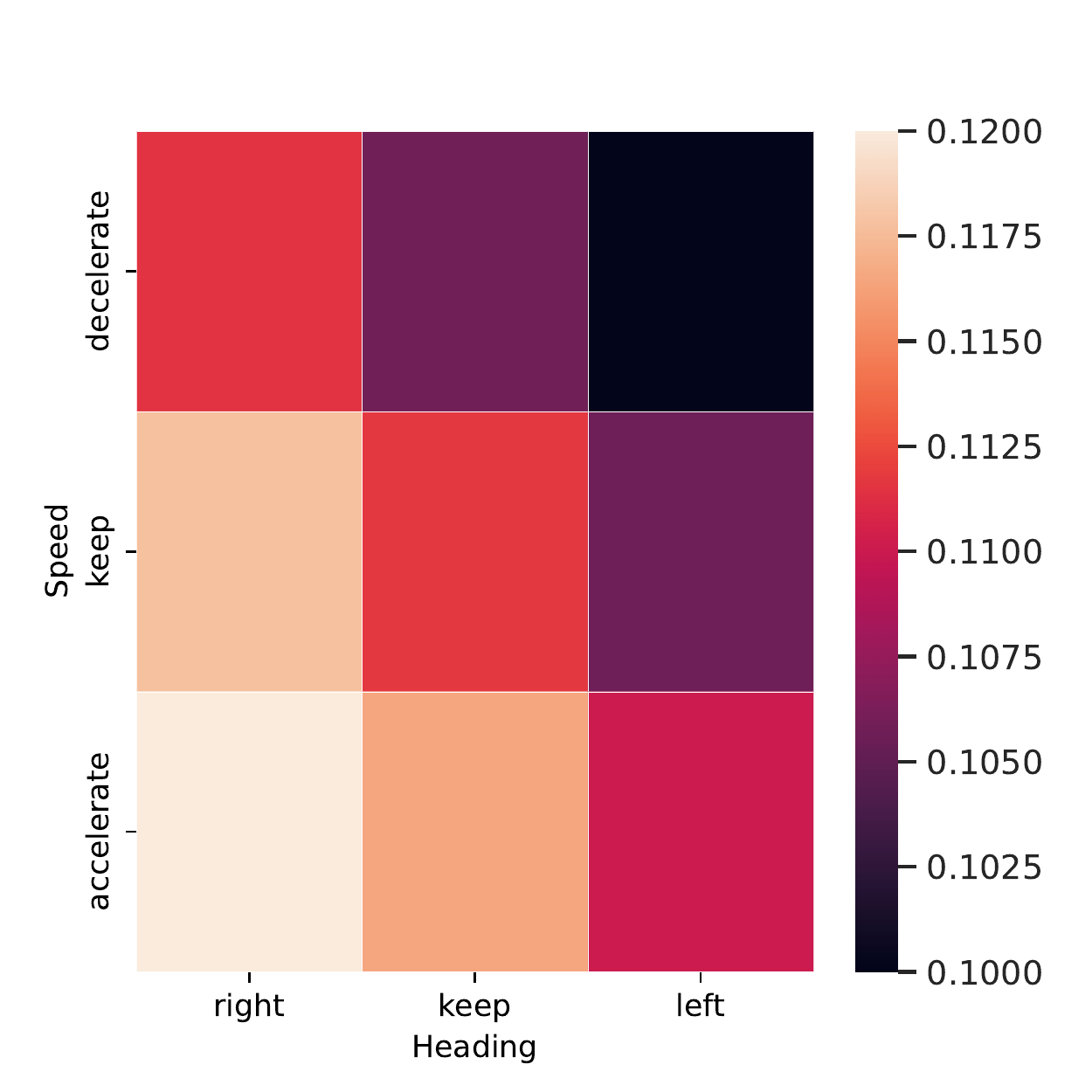}}
\subfigure[The safety action probability map under the safe state]{
\label{Fig.safe.heatmap.SafeDQN}
\includegraphics[width=0.23\linewidth]{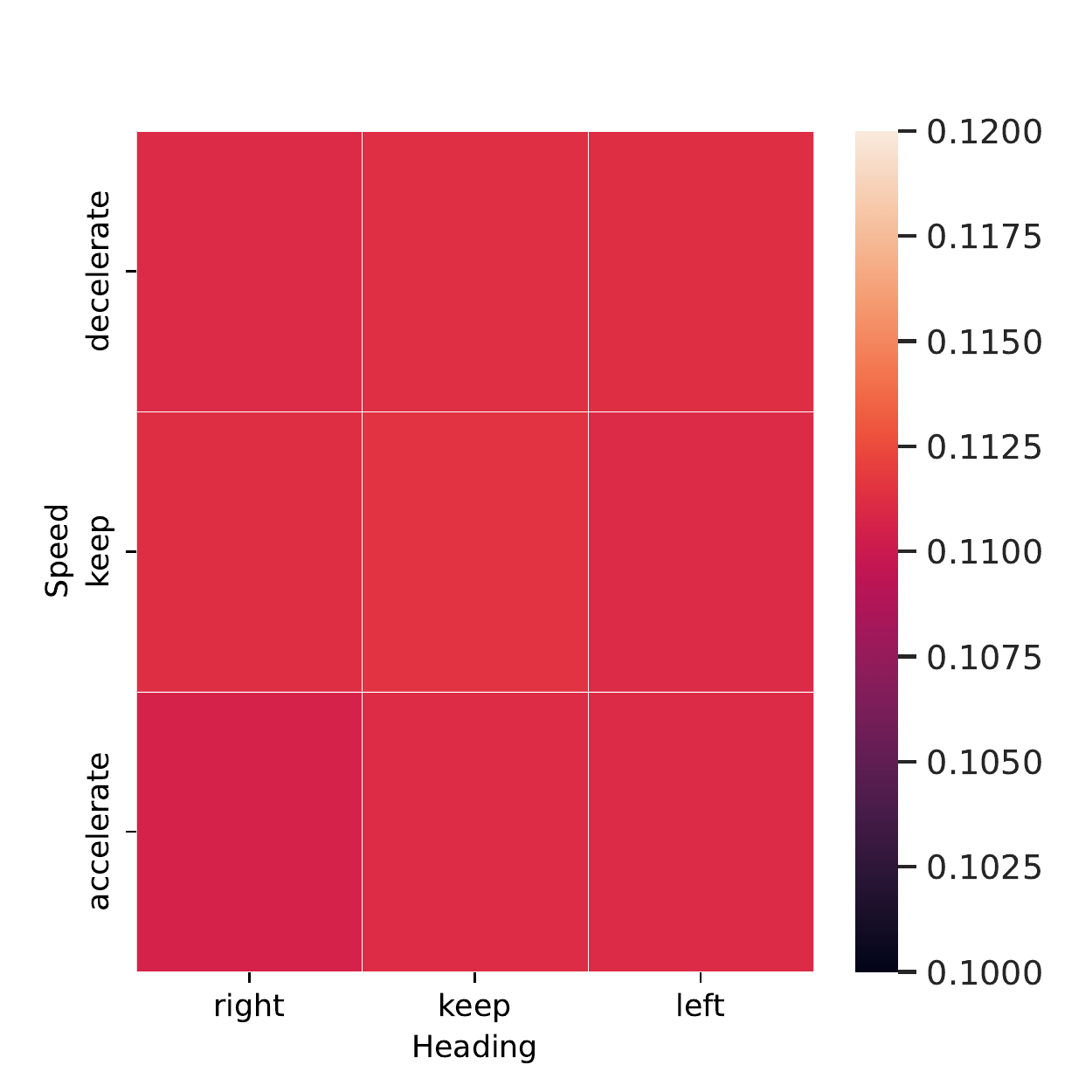}}
\subfigure[A dangerous state]{
\label{Fig.danger.state}
\includegraphics[width=0.23\linewidth]{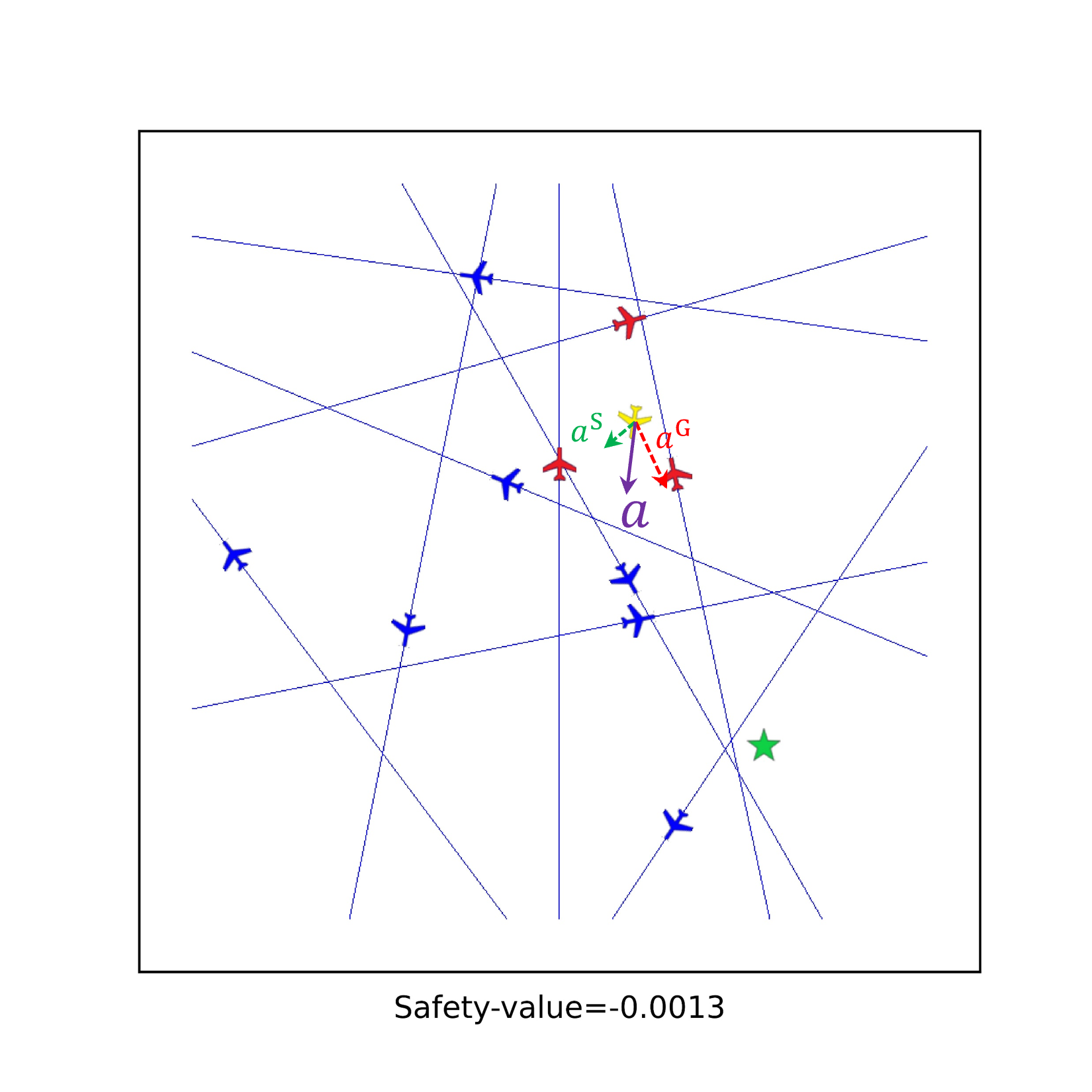}}
\subfigure[The overall action probability map under the dangerous state]{
\label{Fig.danger.heatmap.SafeDQN-X}
\includegraphics[width=0.23\linewidth]{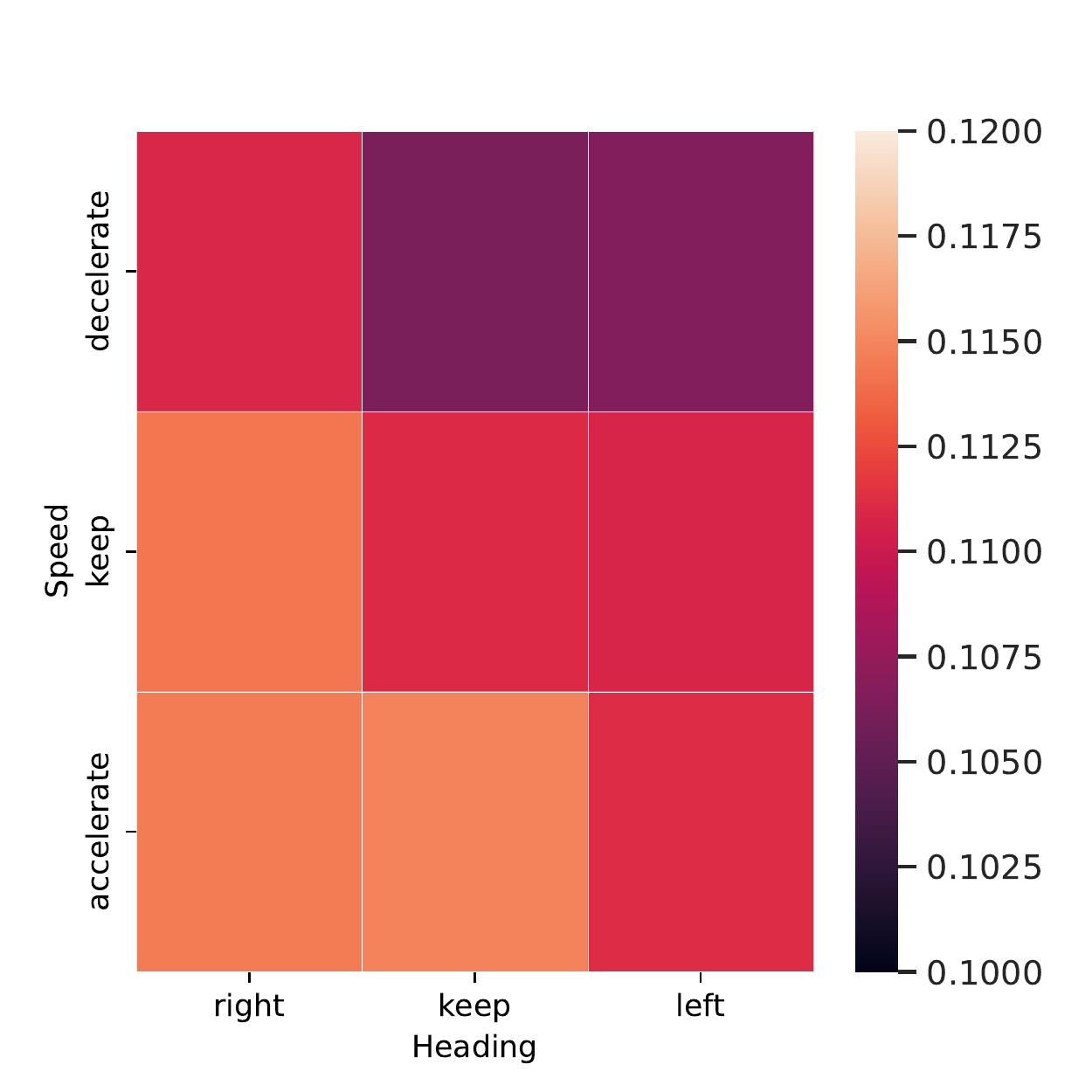}}
\subfigure[The goal action probability map under the dangerous state]{
\label{Fig.danger.heatmap.GoalDQN}
\includegraphics[width=0.23\linewidth]{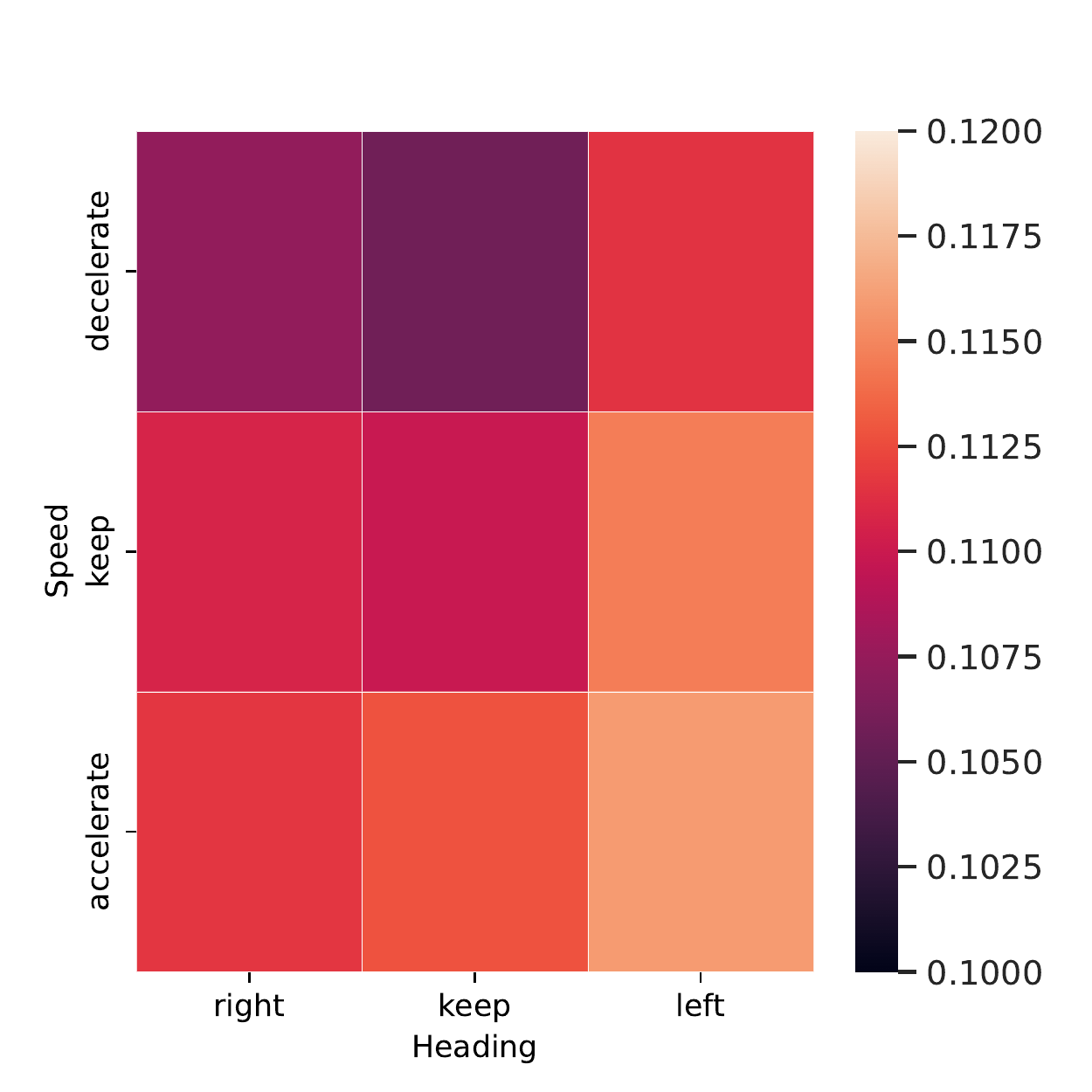}}
\subfigure[The safety action probability map under the dangerous state]{
\label{Fig.danger.heatmap.SafeDQN}
\includegraphics[width=0.23\linewidth]{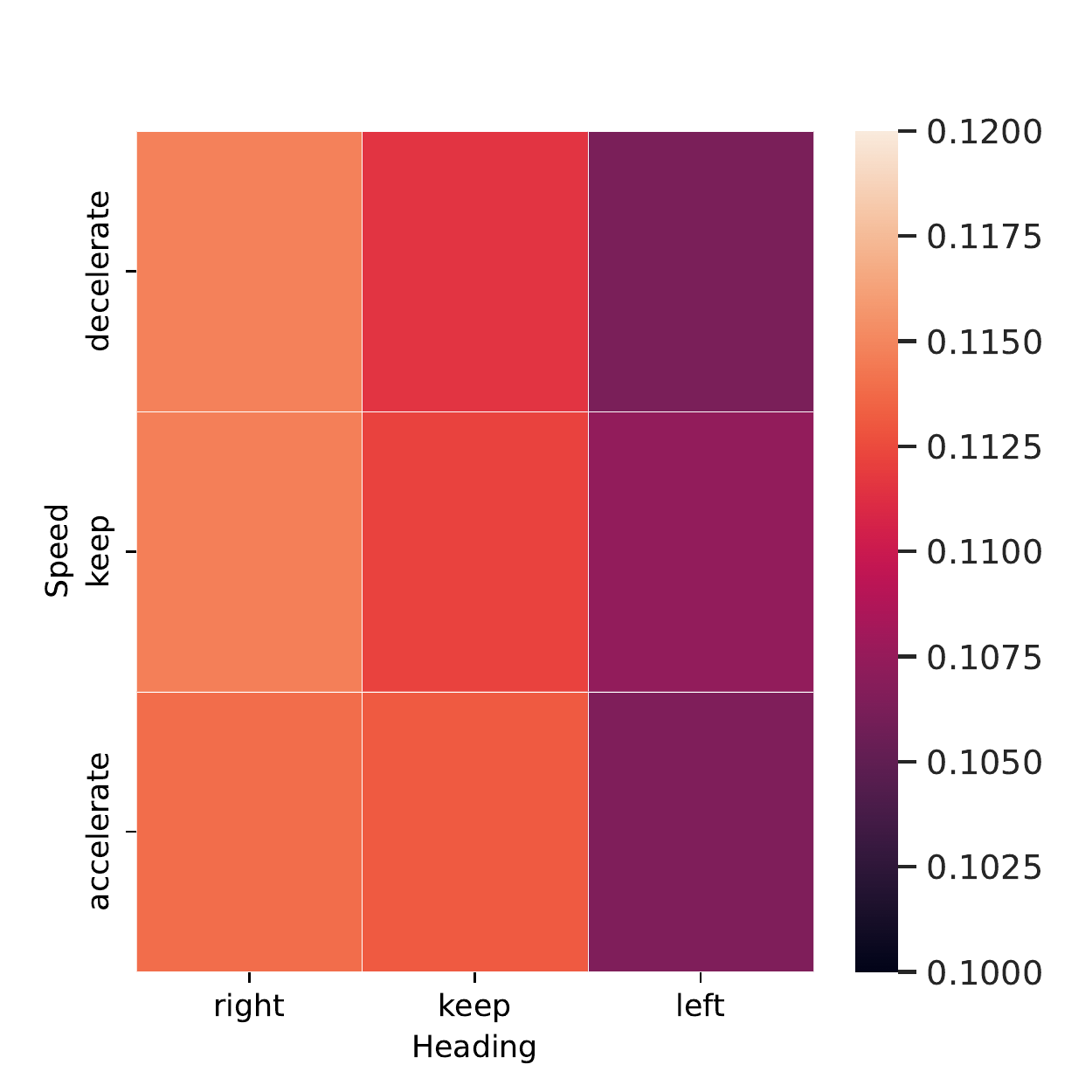}}
\caption{(a) Example of a safe state. The safety-value of the current state determined by Safe DQN is 0.1352. (b)-(d) Overall, goal and safety action heat maps under the safety state. (e) Example of a dangerous state. The safety-value of the current state determined by Safe DQN is ${-0.0013}$. ($a$ marked by purple: the final action, $a^G$ marked by red: the preferred action of Goal DQN, $a^S$ marked by green: the preferred action of Safe DQN) (b)-(d) Overall, goal and safety action heat maps under the dangerous state. The probability map is computed by the corresponding Q values using Equation~\eqref{eq:att_time}. }
\label{SafeDQN-X action visualization 1}
\end{figure*}

In order to test whether the proposed SafeDQN-X can let human users understand the learned behavior, we visualize the safety values and Q values of a set of environments. Two examples are shown in Figure~\ref{SafeDQN-X action visualization 1}. In Figure~\ref{Fig.safe.state}, the free flight is relatively far away from other aircraft, and there is no potential conflict. The safety value $V(\mathbf{x})$ in Equation~\eqref{eq:value_duel} of the current state determined by Safe DQN is 0.1352. In Figure~\ref{Fig.danger.state},  the free flight is close to other aircraft, and it is prone to conflict and collision. The safety value of the current state determined by Safe DQN is $-0.0013$. Clearly, the safety state-value learned by dueling DQN provides an accurate metric measuring the environment's safety. Figure~\ref{Fig.safe.heatmap.SafeDQN-X},\ref{Fig.safe.heatmap.GoalDQN}, \ref{Fig.safe.heatmap.SafeDQN} illustrate the probabilities of different actions given by SafeDQN-X and its two components under the safe state. The probabilities of all actions provided by Safe DQN are similar in the safe state. This is reasonable since any actions taken in this state will not lead to conflicts. In this case, the action of the free flight is mainly determined by the Goal DQN, and the decision is to accelerate and turn right. Figure~\ref{Fig.danger.heatmap.SafeDQN-X},\ref{Fig.danger.heatmap.GoalDQN},\ref{Fig.danger.heatmap.SafeDQN} illustrate the action probabilities under dangerous state. In this case, the action that Goal DQN decides to take is to turn left and accelerate because it can get closer to the target faster. However, the safety value of this action is very low. Therefore, Safe DQN prefers to turn right, making the free flight far away from the target. After combining the two components, the final action is acceleration and keeping the head angle. Those examples verify the 
explainability of our model.

\section{Conclusion and discussion}
\label{sec:con}

We have proposed SafeDQN-X, a prototype for conflict resolution to free flight controllers. We deconstructed the traditional DQN models into two parts, a Safe DQN, perceiving the safety situation of the environments, and a goal DQN, solely aiming to improve efficiency. In addition, SafeDQN-X works by the local nearest neighbors' information, eliminating abundant information from intruders without collision potential. The success of SafeDQN-X comes from a fully explainable design. We showed that, by safe and goal value deconstruction, SafeDQN-X could outperform traditional DQNs in terms of airspace safety. The behavior of SafeDQN-X is also transparent for the explanation, and especially the safety value function provides a direct safety measure of the environment. We also highlighted issues with adversarial attacks on which a strategically-timed attack method is developed. The proposed attack method relies on the deconstruction design of SafeDQN-X and can impose safety-oriented and goal-oriented attacks. A severe issue reported in the experiments is that SafeDQN-X is vulnerable to the specifically designed adversarial. The experiments indicated that the leakage of structure design could lead to severe safety issues. Thus we suggest a principle on "AI-based Controller" that can play an important role in developing intelligent and smart systems in various real-world applications. {\textbf{The AI algorithm should be transparent to explanation but encapsulated to usage.}}

It should be noted that the model presented in this paper is for proof-of-concept studies. The possibilities for future research from this paper are extensive. From the perspective of air traffic control for free flight, ongoing work involves evaluations of more realistic simulation environments. For the neural network structure, the safety and goal value decomposition of more advanced multi-agent reinforcement learning schemes would be desirable for more complex applications. The adversarial attack issues also require further explorations. 

 \section{Acknowledgements}

This work was supported by National Natural Science Foundation of China (Grant No.U20A20161, No. 72201184), the Open Fund of Key Laboratory of Flight Techniques and Flight Safety, Civil Aviation Administration of China (Grant No. FZ2021KF04), and the Fundamental Research Funds for the Central Universities.

\ifCLASSOPTIONcaptionsoff
  \newpage
\fi

\bibliographystyle{IEEEtran}
\bibliography{IEEEfull,mybibfile}


%


%





\begin{IEEEbiography}[{\includegraphics[width=1in,height=1.25in,clip,keepaspectratio]{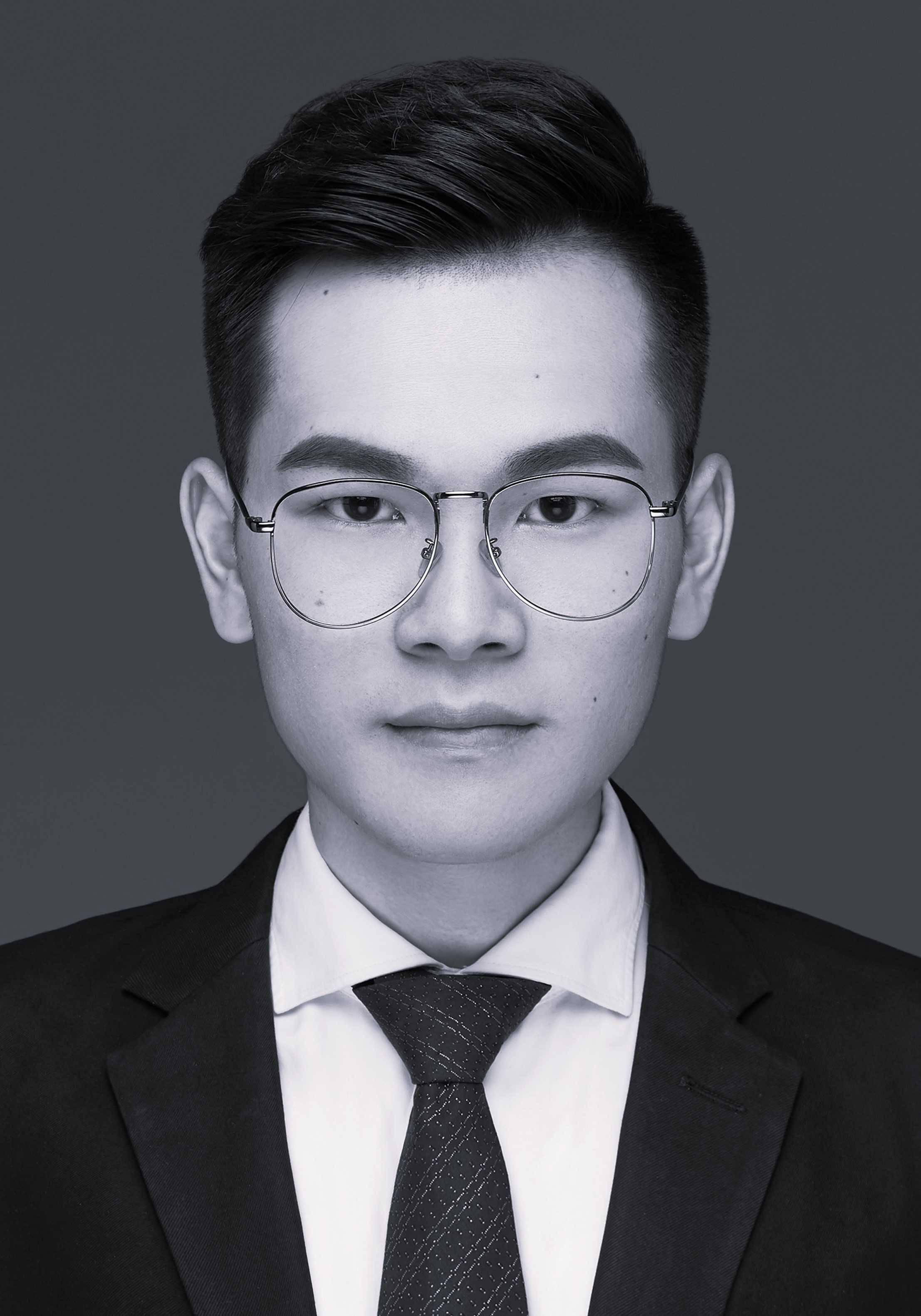}}]
{Lei Wang} received the B.Eng degree in intelligence science and technology from University of Science and Technology Beijing (USTB), Beijing, China, in 2021. Currently, he is a 2nd year Ph.D. student of software engineering, at the College of Computer Science, Sichuan University, China. His research directions include adversairal attack and defense for deep reinforcement learning, and applications to urban air mobility.
\end{IEEEbiography}

\begin{IEEEbiography}[{\includegraphics[width=1in,height=1.25in,clip,keepaspectratio]{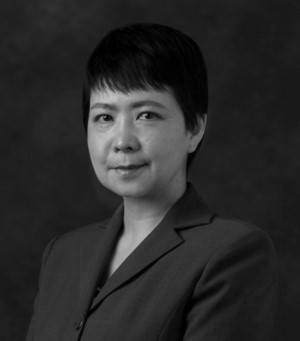}}]
{Hongyu Yang} received B.S. degree from Shanghai Jiaotong University in 1988, and received M.S. and Ph.D. degree from Sichuan University in 1991 and 2008 respectively. She is currently a professor with the College of Computer Science, Sichuan University, China. Her research interests include visual synthesis, digital image processing, computer graphics, real-time software engineering, air traffic information intelligent processing technology, visual navigation, and flight simulation technology.
\end{IEEEbiography}

\begin{IEEEbiography}[{\includegraphics[width=1in,height=1.25in,clip,keepaspectratio]{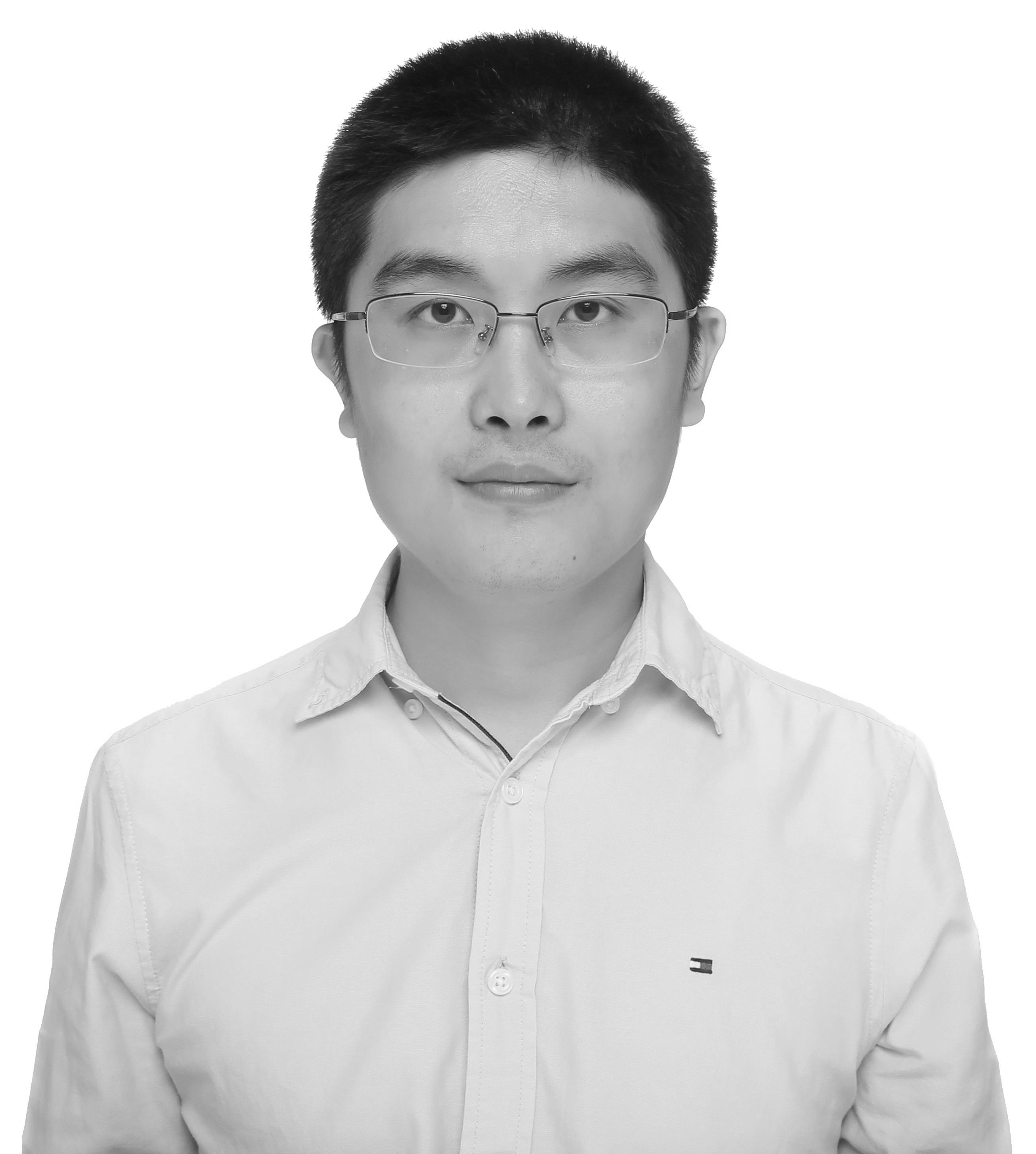}}]
{Yi Lin} (Member, IEEE) received the Ph.D. degree from Sichuan University, Chengdu, China, in 2019. He currently works as an Associate Professor with the College of Computer Science, Sichuan University. He was also a Visiting Scholar with the Department of Civil and Environmental Engineering, University of Wisconsin–Madison, Madison, WI, USA. His research interests include air traffic flow management and planning, machine learning, and deep learning-based air traffic management applications.
\end{IEEEbiography}

\begin{IEEEbiography}[{\includegraphics[width=1in,height=1.25in,clip,keepaspectratio]{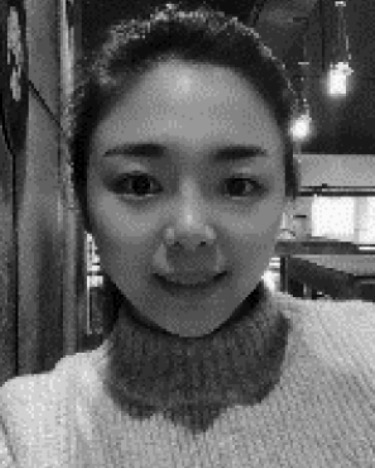}}]
Suwan Yin is an associate professor in ATM/CNS at the College of Computer Science, Sichuan University, China. Prior to joining Sichuan University, she received her Ph.D. in the Department of Civil and Environmental Engineering from Imperial College London. Until now, she has published more than 20 journals (i.e., Transportation Research Part B and C) and conference papers (i.e., AIAA and TRB). Her research interests include integrated airport optimization, airside optimization, network modeling, as well as traffic control and management.
\end{IEEEbiography}

\begin{IEEEbiography}[{\includegraphics[width=1in,height=1.25in,clip,keepaspectratio]{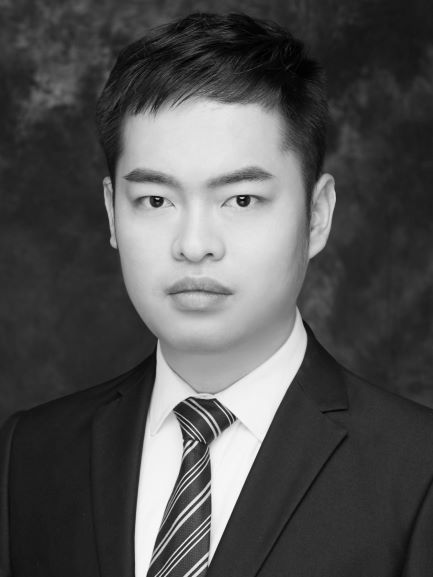}}]
{Yuankai Wu} (Member, IEEE) received the M.S. degree in transportation engineering and the Ph.D. degree in vehicle operation engineering, both from the Beijing Institute of Technology (BIT), Beijing, China, in 2015 and 2019. He is a professor at the College of Computer Science, Sichuan University, China. Prior to joining Sichuan University in March 2022, he was an IVADO postdoc researcher with the Department of Civil Engineering, McGill University. His research interests include intelligent transportation systems, and spatiotemporal data analysis.
\end{IEEEbiography}

\end{document}